\documentclass[preprint,nopreprintline,3p,11pt,authoryear]{elsarticle} 
\usepackage{amssymb}
\usepackage{amsmath}
\usepackage{setspace}
\usepackage[table]{xcolor}
\usepackage{hyperref}
\usepackage{booktabs}
\usepackage{arydshln}
\usepackage{multirow}
\usepackage{bm}
\usepackage{adjustbox}
\usepackage{physics}
\usepackage{graphicx}
\usepackage{colortbl}
\usepackage{subcaption}
\usepackage{enumitem}
\definecolor{lightgray}{gray}{0.9}
\journal{}

\begin{document}

\begin{frontmatter}

\title{Stabilizing distribution-free probabilistic forecasts} 

\author[affil1]{Jente Van Belle\corref{cor1}}
\ead{jente.vanbelle@kuleuven.be}
\author[affil2,affil3]{Honglin Wen}
\ead{linlin00@sjtu.edu.cn}
\author[affil1]{Wouter Verbeke}
\ead{wouter.verbeke@kuleuven.be}
\author[affil3,affil4,affil5,affil6]{Pierre Pinson}
\ead{p.pinson@imperial.ac.uk}
\cortext[cor1]{Corresponding author}
\affiliation[affil1]{organization={Faculty of Economics and Business, KU Leuven},
            country={Belgium}}
\affiliation[affil2]{organization={Department of Electrical Engineering, Shanghai Jiao Tong University},
            country={China}}
\affiliation[affil3]{organization={Dyson School of Design Engineering, Imperial College London},
            country={United Kingdom}}
\affiliation[affil4]{organization={Halfspace},
            country={Denmark}}
\affiliation[affil5]{organization={Department of Technology, Management and Economics, Technical University of Denmark},
            country={Denmark}}
\affiliation[affil6]{organization={CoRE, Aarhus University},
            country={Denmark}}

\begin{abstract}
Multi-step-ahead forecasts are often updated as new observations become available, since shorter forecast horizons typically improve forecast quality.
However, such improvements come at the cost of 
forecast instability, i.e., variability in forecasts for the same target period.
This instability can trigger costly changes to plans formulated based on the forecasts and may erode trust in the forecasting system.
In this work, we integrate forecast stability alongside forecast quality into the 
training of distribution-free probabilistic time-series forecasting models,
allowing us to control this trade-off.
We propose a 
method for generating stabilized forecasted conditional quantile functions using regression splines parameterized by a neural network.
This approach enables joint optimization of quality and stability,
as it allows us to directly penalize 
dissimilarities arising from forecast updates.
Furthermore, it allows assigning 
varying importance to stabilizing different parts of the forecast distributions (e.g., central parts vs.\ tails) to focus on the parts most relevant for the intended downstream use (e.g., the upper tail for inventory management).
We empirically evaluate the proposed method
on two datasets with different statistical properties
and show that it can effectively reduce forecast instability without 
a substantial loss in forecast quality,
and that it can target stabilization effort toward specific parts of the forecast distributions.
\end{abstract}

\begin{keyword}
Forecasting 
\sep Deep learning 
\sep Global models
\sep Splines 
\sep Wasserstein distance
\end{keyword}

\end{frontmatter}

\section{Introduction}
\label{sec:introduction}

Forecasts are typically a means to an end, serving as input to decision-making.
For instance, weather forecasts play a critical role in managing renewable energy production, enabling energy producers to optimize a wide range of operational and financial decisions \citep{sweeney2020renewables}.
Similarly, demand forecasts are crucial for supply chain management, guiding inventory positions, among other aspects
\citep{syntetos2016scforecasting}.
In both cases, forecasts are used as input to (multiple) decision-making processes and are essential for planning operations.
The resulting plans, however, are often not static; they are typically revised 
as new information becomes available.
While it is generally expected that more recent information provides more precise inputs for the decision-making processes,
leading to improved 
decisions and plans,
substantial adjustments to these plans can entail costs associated with (partially) revising or reversing earlier decisions and actions
\citep{terwiesch2005volatility,pappenberger2011jumpiness,tunc2013nervousness}.

In this work, we focus 
specifically 
on updated 
or revised 
forecasts as sources of new information.
Updating forecasts introduces variability in the forecast for a specific time period as time progresses, 
that is, as the forecasting origin (i.e., the time period from which forecasts are generated) moves closer to the target period and new observations become available.
This type of instability has been referred to
by various names,
including 
\textit{forecast churn} \citep{krishnan2007churn},
\textit{forecast inconsistency} or \textit{jumpiness} \citep{zsoter2009jumpiness},
\textit{rolling origin forecast instability} \citep{vanbelle2023stability}, \textit{forecast incongruence} \citep{pritularga2024congruence}, and \textit{vertical 
instability} \citep{godahewa2025stability}.

While an updated forecast is generally of better quality on average due to the shorter forecast horizon, as discussed above, the resulting instability can impact decision-making and entail costs.
These costs are often difficult to quantify and are frequently linked to a decline in trust in the forecasting system or pipeline.
For instance, in 
macroeconomic forecasting, (excess) instability in forecasts may harm the credibility of the issuing institution and prompt revisions to investment plans and strategies.
Similarly, in the case of flood forecasts used to issue warnings or trigger evacuation plans, unstable forecasts can undermine trust and discourage individuals from taking necessary safety precautions.
Furthermore, forecast instability may prompt unwarranted judgmental adjustments, as noted by \citet{nordhaus1987forecasting}: "Moreover, some forecasters might smooth their forecasts as a service to customers. One forecaster told me that he smoothed his forecasts because a more accurate but jumpy forecast would drive his customers crazy" (p.\ 673).
Such (small) judgmental adjustments to algorithmically generated forecasts are known to often degrade forecast quality, 
except when they account for the consequences of extreme events
\citep[see, e.g.,][]{franses2010judgmental}.
Hence, in most contexts, there exists an inherent trade-off between forecast stability and forecast accuracy: if forecasts are not updated, we avoid the costs resulting from forecast instability; however, we also forgo potential improvements in forecast quality.

In Section~\ref{sec:toy_example}, we present a toy example to illustrate that stable and unstable forecasters can also produce qualitatively similar forecasts, making them indistinguishable 
based on
forecast quality alone. 
This example highlights
that forecast stability is often an important additional criterion
to consider alongside forecast quality
when selecting among alternative forecasting models or pipelines, or when optimizing them.
In this paper, we focus on integrating forecast stability directly into the optimization of a global forecasting model,
i.e., a model whose parameters are estimated jointly across multiple time series \citep{januschowski2020criteria}.
Previous research has shown that using a composite loss function to optimize a global neural forecasting model for both forecast quality and stability can enhance the stability of point \citep{vanbelle2023stability} and probabilistic \citep{vanbelle2024probabilistic} forecasts without considerably compromising forecast quality.
The existing method for probabilistic forecasts by \citet{vanbelle2024probabilistic}
focuses on
stabilizing Gaussian output distributions.
Under this assumption,
well-known distributional dissimilarity measures
that can be used to quantify differences between forecasts for the same target period made at different origins
admit closed-form solutions in terms of the distributional parameters, i.e., the means and variances, which correspond to the network outputs.
Because these measures are available in closed form, they can be directly integrated into the loss function to minimize forecast instabilities.
For non-Gaussian data, a similar approach could be adopted for specific parametric families for which closed-form solutions 
to distributional dissimilarity measures exist,
or alternatively, 
the Gaussian approach could 
be combined with a data transformation
such as the 
Yeo-Johnson 
transformation 
\citep{yeo2000}.
However, the prevalence of intermittent time series in industrial settings
\citep{nikolopoulos2021intermittent},
as exemplified by the M5 competition dataset \citep{makridakis2022m5},
highlights the need for a more general approach that can directly produce stabilized full-density forecasts for a variety of output distributions without
requiring explicit 
distributional assumptions.
In this distribution-free setting,
closed-form expressions of forecast instability in terms of distributional parameters are 
necessarily unavailable,
raising the question of which distribution-free forecasting approaches provide the flexibility to integrate a stabilization mechanism directly into model optimization.

Our contribution in this work is to address this gap in the literature by proposing a forecasting method, named StableSQF,
that enables the generation of stabilized distribution-free probabilistic forecasts.
More specifically,
we introduce a method inspired by \citet{gasthaus2019spline} to generate stabilized forecasted conditional quantile functions, modeled using linear isotonic
(i.e., monotonically increasing) 
regression splines,
with the spline parameters partially obtained by training a neural network
that optimizes a discretized approximation of the continuous ranked probability score (CRPS)
\citep[see, e.g.,][]{gneiting2007scoringrules}.
The adopted quantile function approach can flexibly generate a wide range of continuous
output 
distributions 
without requiring a predefined specification.
Beyond this inherent feature, it offers two additional advantages. First, unlike methods that forecast a fixed preselected set of quantiles \citep[e.g.,][]{wen2017multiQRNN,lim2021tft},
it prevents quantile crossing (i.e., the situation in which lower quantiles of a forecast distribution exceed higher ones).
Second, and most importantly for our work, it provides the 
flexibility to optimize forecast stability alongside quality. 
Specifically, it enables us to directly penalize dissimilarities arising from forecast updates, 
tune the relative weight of these penalties to balance forecast quality and stability,
and assign different levels of importance to stabilizing different parts of the forecast distributions.

The ability to control where stability is induced has practical value.
In certain contexts, improving the stability of tail forecasts (potentially at the cost of a slight
loss in quality) may be highly valuable, whereas stability of the center of the distributions is less relevant; in other settings, the opposite may hold.
For instance, when demand forecasts are used as inputs for inventory management, 
specific high quantiles of the product-level forecast distributions may serve as inputs to an optimization engine that drives operational planning, 
while the mean of the product-level forecast distributions is used to compute aggregations across multiple products and over time 
to inform tactical managerial decision-making.
In contrast, in applications such as weather forecasting,
stabilizing the evolution of the central part of forecast distributions may be more important than improving stability in the tails,
as this may help enhance user trust in the forecasts.

Following a discussion of the motivation for considering forecast stability as an additional quality dimension of forecasting models, illustrated by a toy example in Section~\ref{sec:toy_example}, and a review of the literature 
in Section~\ref{sec:related_work},
we present the proposed method for generating stable distribution-free forecasts in Section~\ref{sec:methodology}.
We then empirically examine the trade-off between forecast quality and stability
by applying our method to two datasets with different statistical properties in Section~\ref{sec:experiments},
before concluding with a summary of findings and directions for future research in Section~\ref{sec:conclusion}.

\section{Forecast stability as an additional quality dimension: a motivating example} 
\label{sec:toy_example}

In a supply chain planning context, forecast instabilities arising from forecast updates can lead to `system nervousness', which refers to the need to revise supply plans to realign planned supply with forecasted demand \citep{tunc2013nervousness}.
Quantifying the costs associated with these instabilities is generally challenging, also in this context.
However, to highlight the importance of considering forecast stability alongside forecast quality when evaluating alternative forecasters, we present a toy example in this domain for which these costs can be explicitly quantified.
Specifically, we design a simulation experiment based on a variant of the well-known newsvendor problem.

\subsection{Problem setting}

The newsvendor problem involves a supplier (referred to as the newsvendor) who faces uncertain demand for a perishable product and must determine the inventory level that maximizes expected profit.
We study a setting in which the newsvendor has multiple opportunities to revise the inventory level for the selling period $t$ \citep{deyong2012twoperiod_newsvendor,deyong2018unlimited_newsvendor}.
In our simulation experiment, we assume 
the newsvendor has three ordering opportunities:
\begin{itemize}[itemsep=0pt, parsep=0pt, topsep=0pt, partopsep=0pt]
    \item Initial order at $t-3$: Place 
    order $o_{t-3} \geq 0$ at unit cost $c$ based on the demand forecast $F_{t|t-3}$.    
    \item Intermediate revision at $t-2$: Adjust the initial order based on the updated forecast $F_{t|t-2}$ by ordering additional units $(o_{t-2}>0)$ at increased unit cost $c + c_{e,t-2}$,
    partially or fully canceling the initial order $(o_{t-2}<0)$ at penalty cost $c_{c,t-2}$ per unit, or retaining it $(o_{t-2}=0)$.
    \item Final revision at $t-1$: Make a final adjustment $o_{t-1}$ based on $F_{t|t-1}$, with higher adjustment costs $(c_{e,t-1}>c_{e,t-2})$ and penalties $(c_{c,t-1}>c_{c,t-2})$.
\end{itemize}
At the beginning of period $t$, the total orders $\sum_{i=1}^{3}o_{t|t-i}$ are received, and the actual demand $y_t \sim Y_t$ is observed during period $t$.
Units sold generate revenue $p$ per unit, while unsold units become worthless at the end of the selling period. Any product shortages result in lost revenue.

\subsection{A stable vs.\ unstable forecaster}

\begin{figure}[t!]
    \centering
    \includegraphics[width=\textwidth]{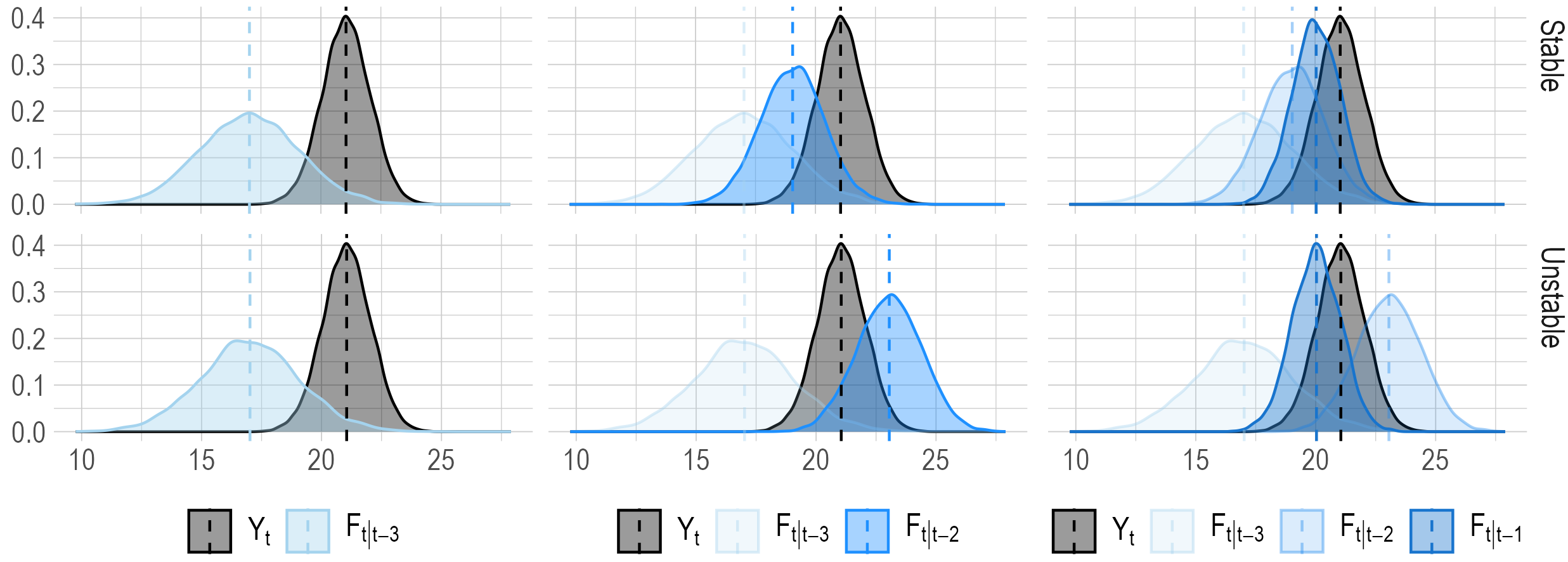}
    \caption{Empirical probability density functions of the ground truth distribution and the forecasts produced by the stable and unstable forecasters for a period $t$, based on 10,000 samples.}
    \label{fig:toy_example_setup}
\end{figure}

Now suppose the newsvendor has access to two alternative forecasters---one stable and one unstable---for making ordering decisions for the selling period $t$ at each time $t-i$, $i=1,2,3$.
The data-generating processes (DGPs) for these forecasters are described in Table~\ref{tab:toy_example_setup}, alongside the ground truth demand distribution.
The forecasts 
are 
modeled as
mixture distributions with two components: the first 
matches the ground truth distribution, indicating that both forecasters have access to 
relevant information, while the second 
represents
distributional bias,
which distinguishes the stable forecaster from the unstable one. 
For both forecasters, we add Gaussian bias that decreases as the target period approaches.
To make the forecast updates of the unstable forecaster more unstable than those of the stable forecaster, we introduce distributional biases with mean shifts that alternate in sign for consecutive updates, whereas for the stable forecaster we maintain a consistent directional shift relative to the ground truth mean.
As a result, 
as visualized in Figure~\ref{fig:toy_example_setup},
the unstable forecaster produces forecast updates that fluctuate around the ground truth distribution, while the stable forecaster generates updates that remain on the same side of the ground truth 
as the initial forecast and gradually converge towards it.

\begin{table}[t!]
    \centering
    \caption{Data-generating processes for the ground truth and forecasters used in the simulation experiment, inspired by the setup in \citet{gneiting2007probabilistic}.}
    \vspace{-0.2cm}
    \begin{adjustbox}{width=\textwidth}
    \begin{tabular}{ccr@{\hskip 0.1cm}c@{\hskip 0.1cm}l}
        \toprule
        \textbf{Forecaster} & \textbf{Origin} & \multicolumn{3}{c}{\textbf{DGP}} \\
        \cmidrule{1-1} \cmidrule(lr){2-2} \cmidrule{3-5}
        Ground truth & $t$ & $Y_t = \mathcal{N}(\mu_t, 1)$; && $\mu_t \sim \mathcal{N}(20, 1)$ \\
        \cmidrule(lr){1-1} \cmidrule(lr){2-2} \cmidrule(lr){3-5}
        \multirow{3}{*}{Stable} & $t-3$ & $F_{t|t-3} = \frac{1}{2}[Y_t + \mathcal{N}(\mu_t + \tau_{t|t-3}, 4)]$; && $\tau_{t|t-3}=\pm 8$ with probability 0.5 each \\
        & $t-2$ & $F_{t|t-2} = \frac{1}{2}[Y_t + \mathcal{N}(\mu_t + \tau_{t|t-2}, 2.5)]$; && \bm{$\tau_{t|t-2} = 
        +\frac{1}{2}\tau_{t|t-3}$} \\
        & $t-1$ & $F_{t|t-1} = \frac{1}{2}[Y_t + \mathcal{N}(\mu_t + \tau_{t|t-1}, 1.75)]$; && \bm{$\tau_{t|t-1} = 
        +\frac{1}{2}\tau_{t|t-2}$} \\
        \cmidrule(lr){1-1} \cmidrule(lr){2-2} \cmidrule(lr){3-5}
        \multirow{3}{*}{Unstable} & $t-3$ & $F_{t|t-3} = \frac{1}{2}[Y_t+\mathcal{N}(\mu_t+\tau_{t|t-3},4)]$; && $\tau_{t|t-3}=\pm 8$ with probability 0.5 each\\
        & $t-2$ & $F_{t|t-2} = \frac{1}{2}[Y_t+\mathcal{N}(\mu_t+\tau_{t|t-2}, 2.5)]$; && \bm{$\tau_{t|t-2}=-\frac{1}{2}\tau_{t|t-3}$} \\
        & $t-1$ & $F_{t|t-1} = \frac{1}{2}[Y_t + \mathcal{N}(\mu_t+\tau_{t|t-1}, 1.75)]$; && \bm{$\tau_{t|t-1}=-\frac{1}{2}\tau_{t|t-2}$} \\
        \bottomrule
    \end{tabular}
    \end{adjustbox}
    \label{tab:toy_example_setup}
\end{table}

\subsection{Evaluating the forecasters}

In Table~\ref{tab:toy_example_forecast_results}, we report the average performance of the forecasters in terms of forecast quality and forecast stability over 10,000 periods.
To assess
quality, we report the average CRPS, with the CRPS for a probability distribution $F$ and an observation $y$ defined as \citep{laio2007}:
\begin{equation} \label{eq:crps}
    \text{CRPS}(F,y) = \int_0^1 \text{QS}_{\alpha}(F^{-1}(\alpha),y) d\alpha,
\end{equation}
where $F^{-1}(\alpha) = \inf\{z \in \mathbb{R}: \alpha \leq F(z)\}$ is the quantile function of $F$, $\alpha \in [0,1]$ is the quantile level, and $F(z)$ is the cumulative distribution function (CDF).\footnote{For other (equivalent) definitions, see, e.g., \citet{gneiting2007scoringrules}.}
The quantile function maps a quantile level $\alpha$ to its corresponding quantile, 
such that the probability that the random variable takes values below this quantile is $\alpha$.
The quantile score $\text{QS}_{\alpha}$ is defined as \citep{gneiting2011}:
\begin{equation} \label{eq:quantile_score}
    \text{QS}_{\alpha}(F^{-1}(\alpha), y) = 2(\mathbb{I}\{y \leq F^{-1}(\alpha)\} - \alpha)(F^{-1}(\alpha) - y),
\end{equation}
where $\mathbb{I}\{y \leq F^{-1}(\alpha)\}$ is the indicator function, which equals 1 if the observation $y$ is less than or equal to the $\alpha$-th quantile of $F$, and 0 otherwise.
The CRPS is a proper scoring rule \citep{gneiting2007scoringrules}, meaning that if an observation is drawn from the distribution $F$, the expected CRPS is minimized when the probabilistic forecast is $F$ rather than any alternative $G\neq F$. 
In the case of deterministic predictions, the CRPS reduces to the absolute error.
To evaluate forecast stability, we report the average 1-Wasserstein ($W_1$) distance between forecasts for the same period made at different origins.
The $p$-Wasserstein distance between two univariate distributions $F$ and $G$ is a function of the difference between their quantile functions \citep{villani2009wasserstein,peyre2019wasserstein}:
\begin{equation} \label{eq:wasserstein}
    W_p(F,G) = \left( \int_0^1 \left| F^{-1}(\alpha) - G^{-1}(\alpha) \right|^p d\alpha \right)^{1/p}.
\end{equation}
For both metrics, we approximate the integrals using discrete sums over 100 quantile levels ($\alpha_k=0.005+0.01k,k=0,1,\dots,99$), where the quantiles are computed empirically based on 10,000 samples per period.

\begin{table}[t!]
    \centering
    \caption{Forecast quality and stability of the forecasters described in Table~\ref{tab:toy_example_setup} evaluated over 10,000 periods, with 10,000 samples generated per period. For both CRPS and $W_1$, lower values indicate better performance.} 
    \vspace{-0.2cm}
    \begin{tabular}{ccccc}
        \toprule
        \textbf{Forecaster} & \textbf{Origin} & \textbf{CRPS} & 
        $\bm{W_1}$ \textbf{adjacent} & $\bm{W_1}$ \textbf{non-adj.} \\
        \cmidrule{1-1} \cmidrule(lr){2-2} \cmidrule{3-5}
        \multirow{3}{*}{Stable} & $t-3$ & 2.91 & -- & -- \\
        & $t-2$ & 1.43 & 2.00 & -- \\
        & $t-1$ & 0.83 & 1.00 & 3.00 \\
        \cmidrule{1-1} \cmidrule(lr){2-2} \cmidrule{3-5}
        \multirow{3}{*}{Unstable} & $t-3$ & 2.91 & -- & -- \\
        & $t-2$ & 1.44 & 6.00 & -- \\
        & $t-1$ & 0.83 & 3.00 & 3.00 \\
        \bottomrule
    \end{tabular}
    \label{tab:toy_example_forecast_results}
\end{table}

The results in Table~\ref{tab:toy_example_forecast_results} indicate that the 
forecasters are indistinguishable in terms of forecast quality, with quality improving on average as the forecast horizon shortens.
In contrast, the $W_1$ adjacent results,
which capture differences between three-step-ahead and two-step-ahead forecasts, and two-step-ahead and one-step-ahead forecasts,
show that the stable forecaster produces substantially more stable forecasts than the unstable one, as intended.
The $W_1$ non-adjacent results,
which compare three-step-ahead and one-step-ahead forecasts,
show no difference in stability, consistent with the experiment's design.
However, these non-adjacent results are not of primary importance in the decision-making context under consideration, since the newsvendor uses all forecast updates, including the two-step-ahead forecasts, to revise inventory levels. Thus, it is the stability of consecutive forecast updates within the entire sequence that matters.

\subsection{Evaluating the decisions}

Table~\ref{tab:toy_example_dm_results} reports the average profit per period for the newsvendor under high and low profit margins,
and with either the unstable or stable forecaster,
across three different strategies:
\begin{itemize}[itemsep=0pt, parsep=0pt, topsep=0pt, partopsep=0pt]
    \item Optimal (myopic):
    Ordering decisions for time $t$ are based on the profit-maximizing quantiles at the different ordering times $t-3$, $t-2$, and $t-1$, respectively:
    $\uparrow F_{t|t-3}^{-1}\left((p-c)/p\right)$, 
    $\uparrow F_{t|t-2}^{-1}\left((p-c-c_{e,t-2})/p\right)$ and $\downarrow F_{t|t-2}^{-1}\left((p-c+c_{c,t-2})/p\right)$,
    and
    $\uparrow F_{t|t-1}^{-1}\left((p-c-c_{e,t-1})/p\right)$ and $\downarrow F_{t|t-1}^{-1}\left((p-\tilde{c}+c_{c,t-1})/p\right)$.
    Quantiles marked with $\uparrow$ and $\downarrow$ are used to decide whether to order additional units or cancel ordered units, respectively, by comparing the relevant quantile to the total units on order, and $\tilde{c}$ denotes the weighted average unit cost of the units on order for time $t$.
    \item Anticipation: 
    To mitigate the risk of committing to excessive orders for time $t$ at $t-3$ and $t-2$ due to overforecasts, 
    median forecasts are used 
    at $t-3$ and $t-2$ to make ordering and cancellation decisions, while 
    the profit-maximizing quantiles 
    are used 
    at $t-1$.
    \item Procrastination: 
    Placing orders for time $t$ 
    is delayed 
    until $t-1$, at which point orders are placed at a higher unit cost using the profit-maximizing $\uparrow$ quantile at $t-1$.
\end{itemize}
The stable forecasts consistently yield higher average profits than the unstable forecasts across all strategies and both profit margins,
except for the procrastination strategy.
Forecast stability has no impact on profit for the procrastination strategy, as the forecasters produce qualitatively similar results for all forecast horizons, including the one-step-ahead forecasts, which are the only inputs used in this strategy.
The same conclusion applies to the proportion of periods in which stable forecasts outperform unstable forecasts in decision-making, with the former leading to higher profits in the majority of periods.

\begin{table}[t!]
    \centering
    \caption{Profitability of the newsvendor for three different strategies, using stable and unstable forecasts as inputs.
    Average profit per period is calculated over 10,000 periods for which forecasts were generated (and results were reported in Table~\ref{tab:toy_example_forecast_results}). 
    The cost structure is defined as follows: $c \sim U(1,2)$, $c_{e,t-2} \sim U(0.2c,0.3c)$, $c_{c,t-2} \sim U(0.3c,0.5c)$, $c_{e,t-1}=2c_{e,t-2}$, and $c_{c,t-1}=2c_{c,t-2}$.}
    \vspace{-0.2cm}
    \begin{tabular}{cccccc}
        \toprule
        \textbf{Profit margin} & \textbf{Strategy} & \textbf{Unstable} $\mathbf{(U)}$ & \textbf{Stable} $\mathbf{(S)}$ & \textbf{$\mathbf{\Delta \%}$} & \textbf{$\mathbf{S>U (\%)}$} \\
        \cmidrule{1-1} \cmidrule(lr){2-2} \cmidrule{3-6}
        \multirow{3}{*}{High $(p=10)$} & Optimal (myopic)& 162.46 & 163.81 & +0.83 & 81.29 \\
         & Anticipation & 162.40 & 164.41 & +1.24 & 76.95 \\
         & Procrastination & 150.67 & 150.67 & +0.00 & 50.36 \\
        \cmidrule{1-1} \cmidrule(lr){2-2} \cmidrule{3-6}
        \multirow{3}{*}{Low $(p=5)$} & Optimal (myopic) & 63.61 & 64.96 & +2.12 & 79.67 \\
         & Anticipation & 63.21 & 65.39 & +3.49 & 76.22 \\
         & Procrastination & 52.26 & 52.26 & +0.00 & 49.90 \\
        \bottomrule
    \end{tabular}
    \label{tab:toy_example_dm_results}
\end{table}

The above decision-making performance evaluation
highlights the critical impact that forecast instability---which may go unnoticed if forecasts are evaluated solely from a forecast quality perspective---can have on operational efficiency 
in dynamic decision-making contexts. 
Quantifying the effects of forecast instability on decision-making performance is often challenging and, when feasible, typically requires expensive simulations. These simulations usually rely on simplifying assumptions, which can weaken the validity of the conclusions drawn from their outputs.
However, evaluating instability at the level of the forecasts themselves,
which are used as inputs to decision-making,
is straightforward and can serve as a valuable proxy.
Therefore, we recommend incorporating forecast stability as an additional evaluation criterion, alongside forecast quality, when selecting or optimizing forecasting models or pipelines.

\section{Literature review}
\label{sec:related_work}

Previous research has explored several ways to incorporate forecast stability into the forecasting process: 
(i) considering it as an additional quality criterion during model selection, 
(ii) combining forecasts to enhance forecast stability, and 
(iii) directly optimizing global forecasting models for both forecast quality and stability.

\subsection{Stability as an additional criterion in model selection}

The earliest attempts to quantify forecast instability can be traced back to the weather forecasting literature.
More specifically, \citet{ruth2009fcs} proposed the Ruth–Glahn Forecast Convergence Score, which measures the number of significant swings across a sequence of forecast revisions, and \citet{ehret2010ci} introduced the Convergence Index, which evaluates whether successive revisions converge toward the eventual observed value without oscillation.  
\citet{lashley2008} 
recommend incorporating such instability metrics into forecast evaluation frameworks alongside traditional forecast quality metrics
in order to strengthen user confidence in numerical weather forecasts, with the expectation that this would lead decision-makers to make decisions on issues affected by potential weather events earlier.

More recently, \citet{pritularga2024congruence} defined a new metric, termed forecast congruence, as $\frac{1}{n} \sum_{t=1}^{n} \sqrt{\mathrm{var}_h (\hat{y}_{t|t-h})}$,
where $\mathrm{var}_h(\cdot)$ represents the variance of forecasts $\hat{y}_{t|t-h}$ for target period $t$
across different forecast horizons $h$.
Their experiments in an inventory management setting, 
using simulated ARIMA-based demand and real demand data from an FMCG manufacturer, 
demonstrate that accuracy and congruence are only weakly correlated and that incorporating congruence into model selection can improve inventory performance.
Specifically, they find that forecasting methods that achieve good congruence while maintaining acceptable accuracy often lead to better downstream decision-making than those that are preferable in terms of accuracy alone.
In the simulated demand setting, they substantiate this by using the congruence of (theoretical) forecasts based on the DGP as a reference point.
Their results indicate that over-congruence (i.e., achieving better congruence than the DGP forecasts) is a desirable property if it does not come at the cost of a substantial decrease in accuracy.

Forecast stability was also considered as an evaluation criterion by \citet{spiliotis2024} in their study on model updating strategy selection (i.e., how frequently forecasting models should be updated).
They assessed the effect of update frequency on forecast accuracy, computational cost, and stability, for both local exponential smoothing models and global LightGBM models.
For exponential smoothing, their results show that forecasts become more stable when the model form is updated less frequently and fewer (critical) parameters are re-estimated.
Combined with the observation that intermediate updating scenarios can deliver comparable or even better accuracy at lower cost, this empirically challenges the theoretical expectation that models need to be specified using the most recent data.
For LightGBM on the product-store-level M5 dataset,
reducing update frequency also maintains accuracy at lower cost; however, in contrast to exponential smoothing, it results in slightly less stable forecasts.
The authors attribute this to LightGBM's nonparametric learning via threshold-based 
splits of the feature space, which are tightly fitted to the training feature distribution and produce an inherently discontinuous prediction function that is highly sensitive to small changes in input values near split boundaries.

\subsection{Enhancing stability through forecast combination}
\label{sec:related_work_forecastcombination}

Similar to the findings of \cite{pritularga2024congruence},
the link between forecast stability and accuracy was also found to be weak by \citet{zsoter2009jumpiness},
who studied forecast instability (termed forecast inconsistency in their work) in the context of numerical weather forecasting.
They investigated whether ensemble (mean) prediction systems improve the consistency between consecutive forecasts for the same target period compared to single forecasting models,
and found positive evidence, 
with the 
improvements in consistency
increasing as the forecast horizon lengthens.

More recently,
\citet{godahewa2025stability} proposed a more direct use of ensembles to stabilize forecasts,
introducing a post-processing method that stabilizes forecasts from any model and offers explicit control over the accuracy-stability trade-off.
Their approach combines newly generated forecasts with the most recent previous forecasts for the same targets (either the original base forecasts or already stabilized forecasts) via weighted averaging. 
This generalizes a simple stabilization approach introduced by \citet{vanbelle2023stability},
which averages all available forecasts for a target period $t+i$ (the new forecast together with older forecasts with longer forecast horizons) to obtain a stabilized forecast $\tilde{y}_{t+i|t}$,
and is used as a baseline method in their work.
\citet{godahewa2025stability} propose two variants with tunable combination weights: (i) partial interpolation with $\tilde{y}_{t+i|t} = (1-w_s) \hat{y}_{t+i|t} + w_s \hat{y}_{t+i|t-1}$, and (ii) full interpolation with $\tilde{y}_{t+i|t} = (1-w_s) \hat{y}_{t+i|t} + w_s \tilde{y}_{t+i|t-1}$, where $0 \leq w_s \leq 1$ 
controls the trade-off between stability and accuracy.
They find that for relatively small values of $w_s$, the stabilized forecasts can be both more stable and more accurate than the base forecasts. 
A key advantage of this post-processing method is its model-agnostic nature,
which also allows it to be used for stabilizing forecasts generated using different forecasting methodologies at different origins,
including both local and global models, as well as judgmentally adjusted or even fully judgmental forecasts.
While earlier works also explored combining forecasts for the same target originating from different forecasting origins \citep{buizza2008comparison,in2022m5}, 
their goal was to improve accuracy rather than explicitly enhance stability.

\subsection{Enhancing stability through global forecasting model optimization}

While stability can be enhanced through forecast combination, it can also be addressed directly during model training.
\citet{vanbelle2023stability} demonstrated that by optimizing a global neural point forecasting model jointly for forecast accuracy and stability, both can be improved.
They achieved this by introducing a composite loss function,
$\mathcal{L}_\text{combined} = \mathcal{L}_\text{error} + \lambda \; \mathcal{L}_\text{instability}$,
where $\mathcal{L}_\text{error}$ optimizes forecast accuracy,
$\mathcal{L}_\text{instability}$ penalizes differences between forecasts for the same target period made at different origins,
and the hyperparameter $\lambda \geq 0$ controls the impact of the latter term on the optimization.
To operationalize this stabilization approach,
they proposed 
\mbox{N-BEATS-S},
an extension of the N-BEATS deep learning model \citep{oreshkin2020nbeats},
and empirically showed that a range of $\lambda$-values result in forecasts that are not only more stable but also maintain, or even improve, forecast accuracy compared to a model with $\lambda = 0$.
Using larger $\lambda$-values than the aforementioned range can further reduce forecast instability, albeit at the cost of accuracy.
The proposed methodology can thus be used to build global neural forecasting models that inherently generate more stable forecasts,
with the modeler having direct control over the weight of the instability penalty.
Building on this work, \citet{caljon2024dlw} demonstrated that stability can be further improved without compromising accuracy by dynamically adjusting $\lambda$ during training using dynamic loss weighting methods.

\citet{vanbelle2024probabilistic} extended this line of research to 
probabilistic forecasting.
Specifically, they proposed N-N-BEATS-S by modifying
the N-BEATS architecture to generate Gaussian distributions as forecasts,
and further extended it along the lines of \citet{vanbelle2023stability} to optimize the Gaussian forecasts for both forecast quality and stability.
For univariate Gaussian distributions, well-known distributional dissimilarity metrics (e.g., the Kullback-Leibler divergence and the 2-Wasserstein distance) admit closed-form solutions in terms of the distributional parameters.
These metrics can be used to quantify forecast instability 
directly from the distributional parameters, which are the outputs of the N-N-BEATS-S network, and can therefore be seamlessly incorporated into the loss function to stabilize the resulting distributional forecasts.
Their results indicate that,
as in the point forecasting setting,
a certain range of $\lambda$-values give rise to more stable 
forecasts without reducing forecast quality.
However, in contrast to what has been observed for point forecasts, no clear improvements in forecast quality were found.
The authors suggest this may stem from 
a 
different impact on forecast quality of inducing stability in variance forecasts compared to mean forecasts.
Specifically, penalizing 
instabilities of full distributional forecasts
may partly counteract the natural increase in variance for longer forecast horizons,
which often occurs to reflect rising uncertainty
due to
a greater likelihood of unexpected changes in underlying patterns and/or a growing influence of undetected patterns such as trends.

In this paper, we build on this last stream of research by generalizing the approach of directly stabilizing forecasts during model training to distribution-free probabilistic forecasting.
More specifically, we propose a distribution-free forecasting method that allows forecast stability to be integrated as an explicit objective during model optimization, alongside forecast quality.
Our approach, introduced in the next section, further provides the flexibility to vary the importance placed on stabilizing different parts of the forecast distributions (e.g., central parts vs.\ tails).

\section{Stabilizing distribution-free probabilistic forecasts}
\label{sec:methodology}

\begin{figure}[tb!]
    \centering
    \includegraphics[width=\textwidth]{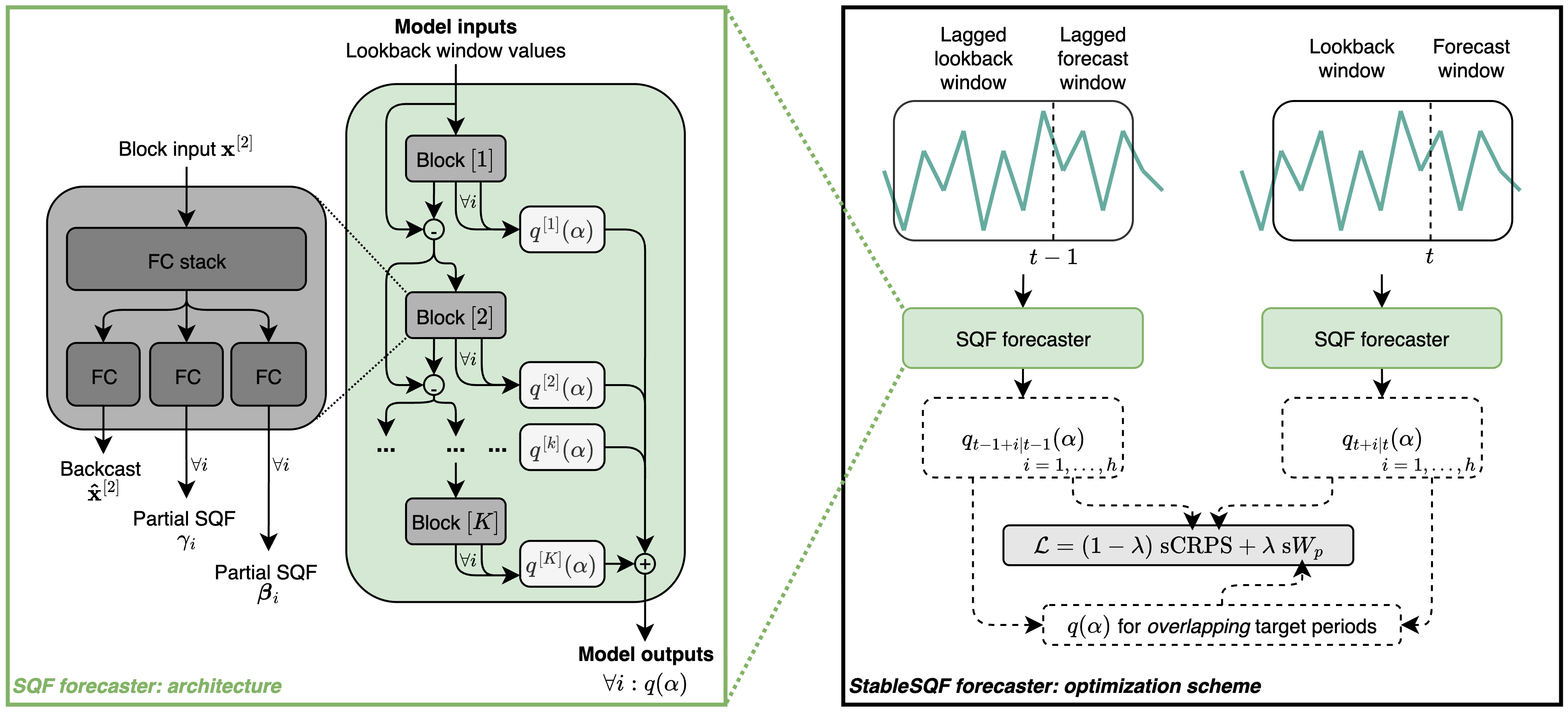}
    \caption{Overview of the SQF forecaster model architecture and the optimization procedure to stabilize the forecasts generated by the SQF forecaster.}
    \label{fig:SQF_and_StableSQF}
\end{figure}

We first introduce the Spline Quantile Function (SQF) forecaster, a flexible neural architecture designed to generate distribution-free probabilistic time-series forecasts by directly modeling full conditional quantile functions.
Next, we present the (architecturally identical) StableSQF forecaster, which uses an augmented optimization procedure to stabilize the generated probabilistic forecasts.
The proposed architecture and optimization procedure are visualized in Figure~\ref{fig:SQF_and_StableSQF}.

\subsection{Spline Quantile Function (SQF) forecaster}
\label{sec:SQF}

Deep learning models for time-series forecasting have seen a surge in adoption and innovation in recent years due to their success when combined with a global learning approach \citep{januschowski2020criteria}
\citep[for a recent overview, see, e.g.,][]{benidis2022dlts}.
This has also led to the development of probabilistic forecasting models.
Initially, two alternative approaches were predominant:
(i) models that generate probabilistic forecasts in the form of a parametric distribution, with network parameters learned via maximum likelihood estimation 
\citep[see, e.g.,][]{salinas2020deepar},
and with the choice of distribution often based on mathematical convenience rather than empirical justification;
and (ii) non-parametric approaches that model only a fixed set of quantiles by optimizing the corresponding quantile scores (see Equation~\eqref{eq:quantile_score})
\citep[see, e.g.,][]{wen2017multiQRNN,lim2021tft}.
The first approach requires a predefined output distribution,
while the second necessitates selecting relevant quantile levels for use during training, 
restricting inference to these quantile levels (i.e., forecasts for unseen quantile levels generally require model re-training),
and can suffer from quantile crossing (i.e., violating the isotonic property of the quantile function).

To address these limitations, one strategy is to use (a sequence of) learnable mappings to transform a simple distribution into a more complex one, leveraging normalizing flows 
\citep[see, e.g.,][]{wen2022nfs}, 
using an Implicit Quantile Network
\citep{gouttes2021iqn},
or modeling the output density as a learned mixture of Gaussians \citep{mukherjee2018armdn}.
Others have tackled these challenges by directly modeling the full conditional quantile function.
Specifically, \citet{gasthaus2019spline} use a spline-based representation of the quantile function, where the spline parameters are the outputs of an 
LSTM-based recurrent neural network (RNN) trained by optimizing the CRPS.
The novelty of their SQF-RNN model lies in the specification of the output distribution, while the RNN architecture itself is similar to those in \citet{wen2017multiQRNN}, \citet{mukherjee2018armdn}, and \citet{salinas2020deepar}.
\citet{park2022iqf} introduce an alternative functional form for representing the quantile function: the Incremental (Spline) Quantile Function (I(S)QF) neural network module.
Their approach models quantiles for a predefined set of training quantile levels by cumulatively adding nonnegative learnable increments
as the quantile levels increase.
For unseen quantile levels, they apply a combination of (learnable) non-parametric linear (spline) interpolation and parametric extrapolation 
strategies for the middle part of the distribution and the tails, respectively.

With the goal of incorporating forecast stability into the optimization of distribution-free probabilistic forecasts,
we adopt the strategy of directly modeling forecasts as full conditional quantile functions.
In doing so, we build on the work of \citet{gasthaus2019spline}
while adopting a direct multi-horizon strategy---also known as the joint or MIMO (multiple input, multiple output) strategy---to generate multi-step-ahead probabilistic forecasts
\citep[as in, e.g.,][]{wen2017multiQRNN,lim2021tft,park2022iqf,vanbelle2024probabilistic},
instead of 
a recursive
modeling approach that produces one-step-ahead forecasts and relies on ancestral sampling for multi-step-ahead predictions \citep{taieb2015multihorizon}.
Having explicit functional forms available at training time that fully characterize the distributions for multiple consecutive target periods---rather than relying on sampling procedures
to generate full probabilistic forecasts \citep[see, e.g.,][]{gouttes2021iqn}---allows us to directly integrate a forecast instability penalty into the loss function, as we will demonstrate in Section~\ref{sec:StableSQF}.

To operationalize the approach outlined above, we introduce the SQF forecaster: a neural architecture that jointly models full conditional quantile functions 
for multiple consecutive target periods.
Specifically, given a lookback window of length $T$
with forecasting origin $t$, 
i.e., a vector of $T$ historical observations
$\vb{x}_{T|t} = [y_{t-T+1},\dots,y_t] \in \mathbb{R}^T$,
the SQF forecaster with model parameters or weights $\boldsymbol{\theta} \in \boldsymbol{\Theta}$ produces $q_{t+i|t}(\alpha|\vb{x}_{T|t};\boldsymbol{\theta})$ as a direct estimate of the conditional quantile function
$F^{-1}_{Y_{t+i}|\vb{x}_{T|t}}(\alpha)$
for each period $t+i$ within a forecast window of length $h$ (i.e., for forecast horizons $i=1,\dots,h$).
The SQF forecaster builds on the \textit{doubly residual stacking} design introduced by \citet{oreshkin2020nbeats} for direct multi-horizon point forecasting.
We adopt this design principle because of its strong empirical performance,
which has been attributed to its ability to sequentially decompose and analyze the raw input signal \citep{oreshkin2021meta}.

Concretely, 
as illustrated in the left panel of Figure~\ref{fig:SQF_and_StableSQF},
the architecture consists of $K$ sequentially connected processing blocks (with $K$ being a hyperparameter),
indexed by $k=1,\dots,K$. 
In each block $k$,
the input $\vb{x}_{T|t}^{[k]}$ is used to generate both a backcast
$\hat{\vb{x}}_{T|t}^{[k]}$
and a forecasted (partial) quantile function
$q_{\gamma,\boldsymbol{\beta}}^{[k]}(\alpha)$
for each period in the forecast window.\footnote{For simplicity,
we omit block-level indices, the conditional formulation, and time indices when clear from context.}
Each partial quantile function is parameterized as a \textit{linear isotonic regression spline} \citep{gasthaus2019spline},
i.e., a piecewise-linear function of the form:
\begin{equation} \label{eq:spline}
    q_{\gamma,\boldsymbol{\beta}}(\alpha)=\gamma+\sum_{l=1}^{L}(\beta_l-\beta_{l-1})(\alpha-d_l)_+,
\end{equation}
where $\gamma \in \mathbb{R}$ is a learnable intercept term,
$\boldsymbol{\beta} \in \mathbb{R}^L_+$ 
are non-negative learnable parameters that determine the slopes of the function pieces (and $\beta_0 = 0$),
$\mathbf{d} \in \mathbb{R}^L$ is a vector of 
ordered knot positions
(i.e., the starting points of the $L$ spline pieces)
satisfying $d_1=0$, $d_L<1$, and $d_l < d_{l+1}$ for $l=1,\dots,L-1$,
and $(x)_+ = \max(x, 0)$.
The slope between any two consecutive knots $d_l$ and $d_{l+1}$ is thus given by 
$b_l=\sum_{l^{\prime}=1}^{l}(\beta_{l^{\prime}}-\beta_{l^{\prime}-1}) = \beta_l \geq 0$,
ensuring that the modeled quantile function is non-decreasing,
as required by its definition.\footnote{Enforcing $b_l \geq 0$ via $b_l=\sum_{l^{\prime}=1}^{l}\beta_{l^{\prime}}$, i.e., replacing $(\beta_l-\beta_{l-1})$ in Equation~\eqref{eq:spline} with $\beta_l$ where $\beta_l \geq 0$, is too restrictive, as it would constrain the spline to be a convex function.}
While \citet{gasthaus2019spline} treat the knot positions $\mathbf{d}$ as learnable parameters (with the number of knots and hence the number of spline pieces $L$ being a hyperparameter),
\citet{park2022iqf} argue that this introduces excessive flexibility for efficiently optimizing quantile functions, 
whose domain is
inherently
$[0, 1]$.
Our preliminary experiments confirmed this finding. 
Therefore, we fix the knot positions and treat their configuration as a hyperparameter
(which also determines 
$L$).
The outputs of each block thus consist of a backcast and spline parameters $\gamma_i$ and $\boldsymbol{\beta}_i$ for each forecast horizon,
which are produced via a stack of four fully connected shared layers followed by an output-specific layer for the backcast and spline parameters,
with the latter followed by linear projections, each corresponding to a specific forecast horizon.
All layers use rectified linear unit (ReLU) activations.
To enforce $\beta_l \geq 0 \; \forall l$,
a final ReLU is applied to the $\boldsymbol{\beta}_i$ outputs.
The difference between the block input $\vb{x}^{[k]}_{T|t}$ and its backcast $\hat{\vb{x}}_{T|t}^{[k]}$ is passed as input to the next block:
\begin{equation}
    \vb{x}^{[k+1]}_{T|t} = \vb{x}^{[k]}_{T|t} - \hat{\vb{x}}_{T|t}^{[k]},
\end{equation}
with $x^{[1]}_{T|t} = x_{T|t}$.
The final forecasted conditional quantile function for each target period in the forecast window is obtained by summing the partial quantile functions across all blocks:
\begin{equation}
    q(\alpha) = \sum_{k=1}^{K} q_{\gamma,\boldsymbol{\beta}}^{[k]}(\alpha).
\end{equation}

To train the SQF forecaster, 
we construct a set of $N$ training instances 
$\{(\vb{x}^{(j)}_{T|t}, \vb{y}^{(j)}_{h|t})\}_{j=1}^N$,
where the length $T$ of the lookback windows $\vb{x}^{(j)}_{T|t}$ is set as a multiple of the forecast horizon $h$,
and the forecast windows $\vb{y}^{(j)}_{h|t} \in \mathbb{R}^h = [y^{(j)}_{t+1},\dots,y^{(j)}_{t+h}]$ contain the actual observed values for the $h$ periods immediately following the last time period in the corresponding lookback window.
Training instances are generated by randomly sampling (i) time series from the set of all available series and (ii) forecasting origins (independently for each selected series).\footnote{The index $t$ is used to refer to the forecasting origin, irrespective of the training sample's absolute time period.}
The weights $\boldsymbol{\theta}$ of the SQF forecaster
are estimated by minimizing the average scaled CRPS jointly across all training samples and all target periods $t+i$, $i=1,\dots,h$, for each sample:
\begin{equation} \label{eq:sqf_forecaster_optim}
    \boldsymbol{\theta}^* = \underset{\boldsymbol{\theta} \in \boldsymbol{\Theta}}{\arg\min} \; 
    \frac{1}{Nh}\sum_{j=1}^{N}\sum_{i=1}^{h}\text{sCRPS}^{(j)}_{t,i},
\end{equation}
with
\begin{equation} \label{eq:scrps}    
    \text{sCRPS}^{(j)}_{t,i} = \frac{\text{CRPS}\left(q_{t+i|t}(\cdot\mid\vb{x}_{T|t}^{(j)};\boldsymbol{\theta}), y_{t+i}^{(j)}\right)}
         {\frac{1}{(T-1)} \sum_{s=1}^{T-1}\big|y^{(j)}_{t+1-s}-y^{(j)}_{t-s}\big|},
\end{equation}
where the in-sample mean absolute error (MAE) from the naive forecasting method 
is used to normalize the CRPS,
making it scale-independent so that each training instance contributes equally to the optimization of the network weights.\footnote{We use this scaling procedure, as the CRPS reduces to the absolute error for deterministic predictions.}
Equation~\ref{eq:sqf_forecaster_optim} can be optimized numerically using stochastic gradient descent, with the CRPS integral (see Equation~\ref{eq:crps}) approximated via discretization.

\subsection{StableSQF forecaster}
\label{sec:StableSQF}

Since the SQF forecaster outputs explicit functional forms---in the form of spline-based conditional quantile functions---as distribution-free probabilistic forecasts for multiple consecutive target periods already at training time,
we can adopt the methodology proposed by \citet{vanbelle2023stability} to improve the stability of point forecasts from global direct multi-horizon neural models
(which was extended by \citet{vanbelle2024probabilistic} for Gaussian probabilistic forecasts; see Section~\ref{sec:related_work}).
We now leverage this approach to augment the optimization problem from Equation~\eqref{eq:sqf_forecaster_optim} in order to optimize the SQF forecaster's outputs for both forecast quality and stability,
with the goal of producing inherently more stable distribution-free probabilistic forecasts.

Specifically, as visualized in the right panel of Figure~\ref{fig:SQF_and_StableSQF},
simultaneously optimizing, for each sample $j$, the conditional quantile functions for both a lookback window $\vb{x}_{T|t}^{(j)}$ and a lagged lookback window $\vb{x}_{T|t-1}^{(j)}$,
with the forecasting origin shifted one period backward in time, allows us to add a forecast instability component to the loss function that directly penalizes forecast instability.
This is possible because our choice to model distribution-free probabilistic forecasts in the form of quantile functions enables us to quantify dissimilarities between forecasts made at different origins for the same overlapping target periods in the corresponding forecast and lagged forecast windows, i.e., for periods $t+i$ with $i=1,\dots,h-1$, using the quantile-function-based formulation of the $p$-Wasserstein distance between two univariate distributions, 
as defined in Equation~\eqref{eq:wasserstein}.

Formally, we augment the optimization problem from Equation~\eqref{eq:sqf_forecaster_optim} by using a composite loss function that jointly optimizes forecast quality and stability,
resulting in the StableSQF forecaster:
\begin{equation} \label{eq:stable_sqf_forecaster_optim}
    \boldsymbol{\theta}^* = 
    \underset{\boldsymbol{\theta} \in \boldsymbol{\Theta}}{\arg\min} \; 
    \frac{1}{N}\sum_{j=1}^{N}
    \Biggl[
    \frac{(1-\lambda)}{2h}\sum_{i=1}^{h}    
    \left(\text{sCRPS}_{t,i}^{(j)}+\text{sCRPS}_{t-1,i}^{(j)}\right)
    + \frac{\lambda}{(h-1)}\sum_{i=1}^{h-1}
    \text{s}W_{p,i}^{(j)}
    \Biggr],
\end{equation}
with
\begin{equation} \label{eq:swp}
    \text{s}W_{p,i}^{(j)} = 
    \frac{W_p\left(q_{t+i|t}(\cdot\mid\vb{x}_{T|t}^{(j)};\boldsymbol{\theta}),q_{t+i|t-1}(\cdot\mid\vb{x}_{T|t-1}^{(j)};\boldsymbol{\theta})\right)}
         {\left( \frac{1}{(T-1)} \sum_{s=1}^{T-1}\big|y^{(j)}_{t+1-s}-y^{(j)}_{t-s}\big|^{p} \right)^{1/p}},
\end{equation}
and $\lambda \in [0,1]$ a hyperparameter controlling
the weight assigned to penalizing forecast instability during training.
In line with how the CRPS is normalized and calculated, 
the in-sample root mean $p$-th power error (which reduces to the root mean square error (RMSE) for $p=2$) from the naive forecasting method
is used to normalize the $p$-Wasserstein distances, and the integral in the $p$-Wasserstein distance
(see Equation~\ref{eq:wasserstein})
is approximated via discretization.

An additional benefit of the 
spline-based conditional quantile function approach to characterizing probabilistic forecasts is that the quantile-function-based formulation of the $p$-Wasserstein distance for inducing forecast stability naturally provides the flexibility to selectively stabilize specific parts of the forecast distributions.
As noted by \citet{vanbelle2024probabilistic}, penalizing instabilities of full distributional forecasts 
may partly counteract a natural increase in forecast variance for longer forecast horizons (which reflects rising uncertainty),
potentially hindering concurrent improvements in forecast quality.
Nevertheless, specifically stabilizing the tails may be particularly relevant for certain downstream decision-making applications,
in which specific low or high quantiles are critical (e.g., inventory management).
To penalize instability in specific parts of the forecast distributions during training, 
one can use a quantile-weighted variant of the $p$-Wasserstein distance, defined analogously to the quantile-weighted CRPS proposed by \citet{gneiting2011}:
\begin{equation} \label{eq:wasserstein_w}
    W_p(F,G) = \left( \int_0^1 \left| F^{-1}(\alpha) - G^{-1}(\alpha) \right|^p v(\alpha) d\alpha \right)^{1/p},
\end{equation}
where $v$ is a nonnegative weight function defined on the unit interval.
In addition to $v(\alpha) = 1$ (uniform weighting), \citet{gneiting2011} suggest using $v(\alpha) = \alpha(1-\alpha)$ 
to focus on the center of distributions, or $v(\alpha) = (2\alpha-1)^2$ to focus on the tails.

Finally, note that while it is also possible to directly penalize dissimilarities between forecasts from non-adjacent origins by incorporating additional lagged lookback and forecast windows with origins shifted further back in time, 
we focus on penalizing dissimilarities between forecasts from adjacent origins.
This choice is motivated by two elements:
(i) the assumption that, in many practical settings in which forecasts are continuously updated, all forecast updates can impact downstream decisions (as in the toy example from Section~\ref{sec:toy_example}), so it is the stability of consecutive forecast updates that typically matters; and
(ii) that by stabilizing forecasts from adjacent origins, we also indirectly stabilize forecasts from non-adjacent origins.

\section{Experiments}
\label{sec:experiments}

In this section, we present the results of experiments conducted to assess the performance of the proposed StableSQF forecaster.
Specifically, we aim to answer two main questions: 
\begin{itemize}[itemsep=0pt, parsep=0pt, topsep=0pt, partopsep=0pt]
    \item \textbf{Q1}:
    Can the proposed StableSQF forecaster, 
    which directly penalizes forecast instability, effectively be used
    to enhance the stability of distribution-free probabilistic forecasts, and if so, what cost does this entail in terms of loss in forecast quality?
    \item \textbf{Q2}: 
    Does the use of a quantile-weighted Wasserstein distance enable us to target our stabilization effort 
    toward specific parts of the forecast distributions?
\end{itemize}
Before reporting 
the experimental results that address these questions 
in Sections~\ref{sec:results_Q1} and \ref{sec:results_Q2}, respectively, 
we first provide a detailed description of the experimental design,
including the datasets, evaluation schemes, forecasting baselines, training methodology, and hyperparameter settings.

\subsection{Experimental design}

\subsubsection{Datasets} \label{sec:datasets}

Since this work focuses on integrating forecast stability alongside quality into the optimization of \textit{distribution-free} probabilistic time-series forecasts,
we use two publicly available datasets with different statistical properties in our experiments.
Specifically,
we use two M competition datasets:
the M4 monthly dataset \citep{makridakis2020m4},
a random sample of series from a database compiled at the National Technical University of Athens spanning demographic, financial, industrial, macroeconomic, and microeconomic domains;
and the M5 Walmart sales dataset \citep{makridakis2022m5}.
For the latter, we use 
an aggregated version of the original dataset employed by, e.g., \citet{godahewa2025stability}, 
where daily item-level sales are aggregated across all Walmart stores.\footnote{Leading zeros were removed from each time series prior to calculating the summary statistics reported in Table~\ref{tab:datasets}.}
We refer to this version as the M5 items dataset throughout this work.
Dataset characteristics are provided in Table~\ref{tab:datasets}, 
including the mean value ($\mu$), standard deviation ($\sigma$), squared coefficient of variation (CV$^2$, computed based on non-zero observations only), 
and the average interval between non-zero observations (AIBNZO),
all averaged across time series.
The datasets contain only time series with non-negative
values at all time steps.

\begin{table}[t!]
\centering
\caption{Characteristics of the datasets used. We report minimum / median / maximum values where applicable. The forecast horizon and the number of forecasting origins from the original M competitions are shown in parentheses.}
\vspace{-0.2cm}
\adjustbox{max width=\textwidth}{
\begin{tabular}{ccccccccc}
\toprule
\multirow{2}{*}{\textbf{Dataset}} & \textbf{No. of} & \textbf{No. of} & \multirow{2}{*}{$\boldsymbol{h}$} & \multirow{2}{*}{\textbf{Time series length}} & \multirow{2}{*}{$\boldsymbol{\mu}$} & \multirow{2}{*}{$\boldsymbol{\sigma}$} & \multirow{2}{*}{\textbf{CV$^2$}} & \textbf{Avg. interval b/w} \\
& \textbf{series} & \textbf{origins} & & & & & & \textbf{non-zero obs.} \\
\midrule
M4 monthly & 48,000  & 13 (1) & 6 (18)  & 60 / 220 / 2,812    & 1,000 / 3,474 / 10,000 & 5.97 / 825 / 18,697  & 0.00 / 0.08 / 9.80 & 1 \\
M5 items   & 3,049   & 15 (1) & 14 (28) & 442 / 1,937 / 1,969 & 0.36 / 6.35 / 525      & 0.63 / 4.32 / 383    & 0.04 / 0.33 / 6.97 & 1.00 / 1.11 / 4.43 \\ 
\bottomrule
\end{tabular}}
\label{tab:datasets}
\end{table}

The SBC classification\footnote{Developed for categorizing demand time series.} proposed by \citet{syntetos2005intermittent} categorizes time series as smooth, erratic, intermittent, or lumpy based on their
CV$^2$ and AIBNZO statistics.
The 
M4 monthly dataset primarily consists of smooth series (i.e., low CV$^2$ and low AIBNZO), with the remaining 
5.52\%
classified as erratic (low AIBNZO but high CV$^2$).
In contrast, for the M5 items dataset, only 72.61\% of the series are smooth, with the remainder classified as 6.33\% erratic, 16.79\% intermittent (low CV$^2$ but high AIBNZO, i.e., a substantial number of zero observations), and 4.26\% lumpy (both high CV$^2$ and high AIBNZO).

\subsubsection{Evaluation schemes} \label{sec:evaluation}

Since we aim to evaluate not only forecast quality but also forecast stability,
which requires at least two forecasts for the same period from different origins,
we deviate from the single-origin evaluation
schemes used in the M competitions.
We use the original M competition test sets for out-of-sample evaluation (the last 18 data points of each time series for the M4 dataset and the last 28 data points for the M5 dataset).
However, we do not produce one- to 18/28-step-ahead forecasts from a single origin as required in the original competitions.
Instead, we apply a rolling origin evaluation \citep{tashman2000rollingoriginevaluation}.
Specifically, we generate one- to six-month-ahead forecasts for the 
M4 dataset (resulting in 13 forecast traces from 13 different origins)
and one- to 14-day-ahead forecasts for the M5 dataset (resulting in 15 forecast traces from 15 different origins).

To evaluate forecast quality, we use the CRPS (see Equation~\eqref{eq:crps}),
while 
stability is evaluated using the 1-Wasserstein distance (see Equation~\eqref{eq:wasserstein}) between forecasts from adjacent origins for the same periods.
For both CRPS and 1-Wasserstein,
the integrals are approximated 
by summing over 100 quantile levels ($\alpha_k=0.005+0.01k, \; k=0,1,\dots,99$), 
with negative quantile function evaluations replaced by zeros.
To report dataset-level performance, for both metrics, we average across horizons, origins, and time series.
Before averaging, we make the CRPSs and 1-Wasserstein distances scale-independent, as described in Sections~\ref{sec:SQF} and \ref{sec:StableSQF}, 
using the
full history of the time series for scaling
(instead of only the values in the lookback windows; see Equations~\eqref{eq:scrps} and \eqref{eq:swp}),
as is standard practice.
The resulting scaled metrics are denoted sCRPS and s$W_1$.

To address the second question 
(\textbf{Q2}) 
outlined above,
we also report quantile-weighted 1-Wasserstein distances (see Equation~\eqref{eq:wasserstein_w}), denoted s$W_1^\text{c}$ and s$W_1^\text{t}$,
which emphasize the center and tails of the distributions, respectively.
Quantile-weighted versions of the CRPS,
denoted sCRPS$^\text{c}$ and sCRPS$^\text{t}$,
are defined analogously.

\subsubsection{Forecasting baselines}

In addition to examining the trade-off between forecast quality and stability
by reporting results for the proposed SQF (see Section~\ref{sec:SQF}) and StableSQF (see Section~\ref{sec:StableSQF}) forecasters with different values of the instability penalty weight $\lambda$ for the latter (see Equation~\ref{eq:stable_sqf_forecaster_optim}),
we also report results for four groups of baseline models: ETS, mean, snaive, and SQF-stabilized.
The forecasting methods in the first group serve as baselines for forecast quality, 
while those in the second group result in highly stable forecasts by construction.
For forecast horizons $h$ that are less than or equal to the seasonal period (as is the case for the M4 monthly dataset, with $h=6$ and a seasonal period of 12),
the methods in the third group produce (nearly) perfectly stable forecasts by construction.
The models in the fourth group are designed to enhance forecast stability through the application of specific post-processing methods:
\begin{itemize}[itemsep=0pt, parsep=0pt, topsep=0pt, partopsep=0pt]
    \item ETS: Exponential smoothing forecasts 
    using the \texttt{ets} function from the \texttt{forecast R} package \citep{hyndman2008forecastR,hyndman2025forecastR}, which relies on an automated model selection procedure.
    We report results for ETS-G and ETS-B.
    ETS-G produces Gaussian
    forecasts, 
    with quantile forecasts obtained analytically when possible and derived from simulated future sample paths using normally distributed errors otherwise.
    ETS-B uses bootstrapping to obtain quantile forecasts, 
    with 5,000 future sample paths simulated based on resampled errors, thereby relaxing the Gaussian assumption.    
    \item mean: Mean forecasts 
    using the \texttt{meanf} function from the \texttt{forecast R} package, 
    using only the most recent 
    observations in a lookback window 
    of the same length as used for
    the (Stable)SQF models (see Table~\ref{tab:hyperparams}).
    As with ETS, we report results for both mean-G and mean-B.
    \item snaive: Seasonal na\"ive forecasts 
    using the \texttt{snaive} function from the \texttt{forecast R} package.
    As with ETS and mean, we report results for both snaive-G and snaive-B.
    \item SQF-stabilized: A natural way to induce stability in forecasts is to combine them with longer-horizon forecasts from earlier origins for the same target period.
    To stabilize SQF forecasts, we apply the partial (SQF-stabilized-P) and full (SQF-stabilized-F) 
    interpolation post-processing schemes introduced by \citet{godahewa2025stability} for stabilizing point forecasts (see Section~\ref{sec:related_work_forecastcombination}): for each overlapping target period, the SQF forecast from the current origin is averaged with the corresponding raw or stabilized SQF forecast from the immediately preceding origin.
    To translate this procedure to a probabilistic forecasting setting, averaging is performed over the quantile forecasts used to approximate the CRPS and 1-Wasserstein distance.
    We report results for low, medium, high, and maximum stabilization strengths, $w_s = 0.25$, $0.5$, $0.75$, and $1$, denoted by the suffixes lo, med, hi, and max, respectively.
\end{itemize}

\subsubsection{Training methodology}

To train the (Stable)SQF forecasters, theoretically, no data pre-processing is required, as the loss functions used are scale-independent.
However, to promote
optimization stability, we standardize each time series prior to training.\footnote{Means and standard deviations used for standardization are computed excluding test set observations.}
As described in Section~\ref{sec:SQF}, training batches of fixed size
are created from these standardized series by first sampling time series uniformly at random with replacement.
Then, for each selected time series, a time step is sampled from a forecasting origin range of tunable length. 
This range comprises the most recent time steps from which training samples can be generated without introducing missing values,
i.e., time steps for which observations are available for all positions in the corresponding (lagged) lookback and forecast windows (see Figure~\ref{fig:SQF_and_StableSQF})\footnote{Time series are zero-padded at the beginning if they contain too few observations to generate at least one valid training sample,
i.e.,
we prepend zero-valued observations to ensure that each time series is included during training.}.
A larger forecasting origin range compensates for a smaller number of time series in the dataset 
by allowing a greater variety of samples to be drawn from each series.
During training, data augmentation is applied on the fly. 
Specifically, each 
series in a batch is first shifted by adding the same random value from the range $(-1, 1)$ to each observation, and then scaled by multiplying the entire series by a random positive factor sampled from $[0.5, 1.5)$.

We optimize the network parameters using the Adam optimizer \citep{kingma2014adam}
with tuned initial learning rates.
To reduce sensitivity to random initialization,
which affects weight initialization, data augmentation, and batch sequence generation,
and to promote convergence without requiring the learning rate to decay to near zero, 
all
results are based on Exponential Moving Average (EMA) models \citep{moralesbrotons2024ema}.
These models maintain a separate set of weights, $\boldsymbol{\theta}^\text{EMA}$, computed as an EMA of the network parameters
without affecting the optimization itself.
Specifically, let $\boldsymbol{\theta}_l$ denote the weights at learning iteration $l \geq 0$ during stochastic gradient descent; the EMA weights are then recursively computed as
$\boldsymbol{\theta}^\text{EMA}_0 = \boldsymbol{\theta}_0, \; \boldsymbol{\theta}^\text{EMA}_{l+1} = \phi\boldsymbol{\theta}^\text{EMA}_l+(1-\phi)\boldsymbol{\theta}_{l+1}$,
where $\phi \in [0,1]$ is a tunable smoothing coefficient.
EMA models have been shown to improve the  
consistency of training outcomes
across different training runs,
which we 
observed
in our experiments; 
consequently, all results are reported based on a single run per hyperparameter combination.

In this experimental evaluation, for StableSQF models, we focus on s$W_1$ variants to induce forecast stability (and also to evaluate it; see Section~\ref{sec:evaluation}), using (quantile-weighted) 1-Wasserstein distances in Equation~\eqref{eq:stable_sqf_forecaster_optim}.
Although lagged lookback and forecast windows are strictly required only to enable forecast instability quantification during the training of StableSQF models,
we also use them when training SQF models (with $\lambda = 0$) to ensure a fair comparison
by using the same number of training samples.

All neural models are implemented and trained in PyTorch Lightning \citep{Falcon_PyTorch_Lightning_2019},
with our code
available at \url{https://github.com/JenteVB/StableSQF}.

\subsubsection{Hyperparameters}\label{sec:hyperpars}

Table~\ref{tab:hyperparams} provides an overview of the hyperparameter values used for training the SQF networks.
They are selected via grid search based on the minimum validation sCRPS (among runs with converged validation losses) from a single run.
We thus only take into account forecast quality for hyperparameter tuning,
as this represents the natural baseline scenario.
The validation sCRPS is calculated on a holdout set comprising, for each time series, the 18 (for 
M4) or 28 (for M5) data points immediately preceding the first data point in the test set.
The same rolling origin evaluation procedure used for the test set results is also used to assess performance on the validation set.

\begin{table}[t!]
    \centering
    \caption{Hyperparameter values for (Stable)SQF models.}
    \vspace{-0.2cm}
    \begin{tabular}{lcc}
        \toprule
        & \textbf{M4 monthly} & \textbf{M5 items} \\
        \midrule
        Lookback window length $T$ & $8h$ & $26h$ \\
        No. of blocks $K$ & 10 & 5 \\
        Hidden layer width & 512 & 512 \\
        Batch size & 512 & 512 \\
        Forecasting origin range & $10h$ & unrestricted \\
        Learning iterations & 11,500 & 12,500 \\
        Initial learning rate & 1e-3 & 1e-3 \\
        EMA $\phi$ & 0.99 & 0.995 \\
        \bottomrule
    \end{tabular}
    \label{tab:hyperparams}
\end{table}

An additional hyperparameter 
that is 
not included in Table~\ref{tab:hyperparams} is the vector of fixed knot positions
(which also determines the number of
pieces $L$)
used in the splines that model the distributional forecasts.
In our experiments, we use 
$L=30$ with the following knot positions:
$0$, $0.01$, $0.025$, and then increments of $0.025$ up to $0.1$;
from $0.1$ to $0.4$, increments of $0.0375$ are used; between $0.4$ and $0.6$, the increment is $0.05$; 
and from $0.6$ to $0.99$, the knot spacing pattern used between $0.01$ and $0.4$ is mirrored.
In line with \citet{park2022iqf}, we found that using smaller intervals 
(i.e., denser knot spacing)
near the lower and upper quantiles,
compared to the center,
improves performance by allowing
better modeling of the typical higher curvature of the quantile functions in those parts.

For the StableSQF models, we use the same hyperparameter values as for SQF.
The additional hyperparameter $\lambda \in [0,1]$,
which controls the weight of the forecast instability penalty during training, 
is not tuned. 
Instead, we report test set performance for several values of $\lambda$ in terms of forecast quality and stability to assess the trade-off between the two.
Tuning $\lambda$ based on the validation sCRPS would cause stability to be ignored in the absence of Pareto improvements.
However, forecast users may prefer slightly less accurate but more stable forecasts over unstable forecasts with only marginally better average forecast quality.

\subsection{Results}

\subsubsection{Trading off forecast quality and stability with StableSQF} 
\label{sec:results_Q1}

\begin{figure}[t!]
\centering
\begin{subfigure}{0.49\textwidth}
    \centering
    \includegraphics[height=7.5cm]{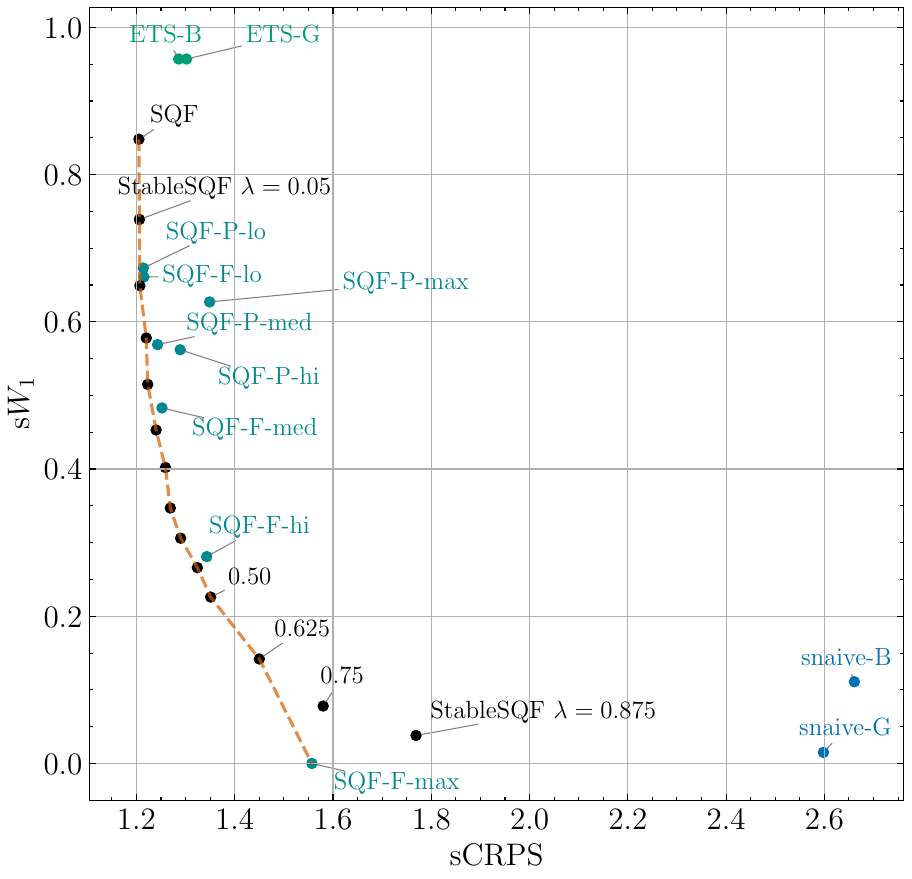}
    \caption{M4 monthly}
    \label{fig:q1_m4}
\end{subfigure}
\hfill
\begin{subfigure}{0.49\textwidth}
    \centering
    \includegraphics[height=7.5cm]{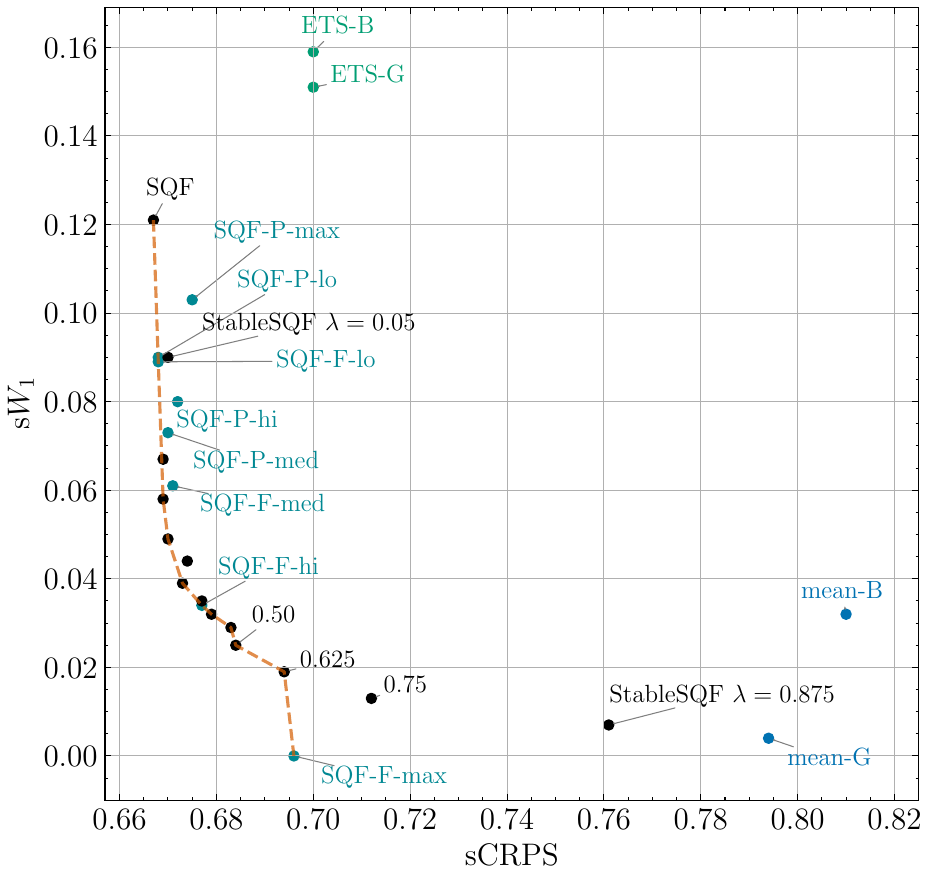}
    \caption{M5 items}
    \label{fig:q1_m5}
\end{subfigure}
\caption{Pareto frontiers of forecast quality (sCRPS) versus 
stability (s$W_1$), with lower values indicating better performance for both metrics.
Each point represents a model, including baseline models (ETS, snaive/mean, and SQF-stabilized variants) as well as StableSQF variants with increasing instability penalty weights 
$\lambda \in [0,1]$ (with SQF corresponding to $\lambda = 0$; StableSQF models with $\lambda \in (0.05, 0.5)$ are left unlabeled for visual clarity). 
The plots illustrate the trade-off between forecast quality and stability: stronger instability penalties improve forecast stability at the cost of forecast quality.
For SQF-stabilized models, the word `stabilized' is omitted in the labels, and only the P/F-stabilization strength suffix is shown to avoid cluttering the plots.}
\label{fig:q1}
\end{figure}

Figures~\ref{fig:q1_m4} and \ref{fig:q1_m5} show the Pareto frontiers of forecast quality (sCRPS) versus forecast stability (sW1) for the M4 monthly and M5 items datasets, respectively.
The frontiers clearly illustrate that StableSQF models can be used to improve forecast stability (lower s$W_1$),
and that this comes at the cost of 
forecast quality (higher sCRPS).
For most $\lambda$ values, the 
StableSQF models lie on the frontier, with the ETS baseline and, for the M4 and M5 datasets respectively, the snaive and mean baselines appearing at opposite ends of the trade-off. 
The former yields less stable but more accurate forecasts, and the latter more stable but less accurate ones. 
However, these baselines do not lie on the Pareto frontier and are therefore dominated by the SQF variants.

For both datasets, small values of $\lambda$ result in only marginal increases in sCRPS while leading to substantial improvements in stability. 
Larger $\lambda$ values $(\geq 0.625)$ yield further improvements in stability but sharply degrade forecast quality,
to the extent that StableSQF models with $\lambda \geq 0.75$ become Pareto dominated by SQF-stabilized-F-max.
Both SQF-stabilized-P and SQF-stabilized-F variants perform well at low stabilization strengths and lie on or near the frontier. 
However, at higher stabilization strengths, SQF-stabilized-P becomes Pareto inefficient: 
it does not yield further improvements in stability but does further reduce forecast quality. 
In contrast, SQF-stabilized-F continues to perform well and ultimately Pareto dominates StableSQF.

\begin{table}[t!]
\centering
\caption{Forecast quality (sCRPS) and 
stability (s$W_1$) for SQF, StableSQF variants, and baseline models (SQF-stabilized, snaive, mean, and ETS variants).
Lower values indicate better performance for both metrics.
Results are reported for selected StableSQF variants with increasing instability penalty weights $\lambda \in [0,1]$ (SQF corresponds to $\lambda = 0$).
For specific relative deteriorations in forecast quality (increases in sCRPS), we report the corresponding relative improvements in forecast stability (decreases in s$W_1$), 
both expressed relative to SQF ($\Delta\%$). 
Metric values for StableSQF models corresponding to 0.5\%, 1\%, 2.5\%, 5\%, and 10\% increases in sCRPS are obtained via interpolation.
These results are based on the same data that underlies Figure~\ref{fig:q1}
and illustrate the quality--stability trade-off.}
\label{tab:q1}
\vspace{-0.2cm}
\resizebox{\textwidth}{!}{
\begin{tabular}{l c ccccc c ccccc}
\toprule
&& \multicolumn{5}{c}{\textbf{M4 monthly}} && \multicolumn{5}{c}{\textbf{M5 items}} \\
\cmidrule(lr){3-7} \cmidrule(lr){9-13}
&& \textbf{sCRPS} & $\bm{\Delta\%}$ 
&& \textbf{s$\bm{W_1}$} & $\bm{\Delta\%}$
&& \textbf{sCRPS} & $\bm{\Delta\%}$
&& \textbf{s$\bm{W_1}$} & $\bm{\Delta\%}$ \\
\midrule
SQF && 1.205 & -- && 0.848 & -- && 0.667 & -- && 0.121 & -- \\
\midrule
StableSQF $\lambda=0.05$ && 1.206 & 0.1 && 0.739 & -12.9 && 0.670 & 0.4 && 0.090 & -25.6 \\
\hdashline
StableSQF $\lambda=0.115/0.211$ && 1.211 & 0.5 && 0.627 &
-26.1 && 0.670 & 0.5 && 0.048 & -60.4 \\
StableSQF $\lambda=0.139/0.267$ && 1.217 & 1.0 && 0.594 & -29.9 && 0.674 & 1.0 && 0.042 & -65.0 \\
StableSQF $\lambda=0.236/0.484$ && 1.235 & 2.5 && 0.471 & -44.5 && 0.684 & 2.5 && 0.026 & -78.3 \\
StableSQF $\lambda=0.331/0.669$ && 1.265 & 5.0 && 0.368 & -56.6 && 0.700 & 5.0 && 0.017 & -86.0 \\
StableSQF $\lambda=0.453/0.805$ && 1.326 & 10.0 && 0.264 & -68.9 && 0.734 & 10.0 && 0.010 & -91.5 \\
\hdashline
StableSQF $\lambda=0.875$ && 1.769 & 46.8 && 0.038 & -95.5 && 0.761 & 14.1 && 0.007 & -94.2 \\
\midrule
SQF-stabilized-P-lo  && 1.214 & 0.7 && 0.673 & -20.6 && 0.668 & 0.1 && 0.090 & -25.6 \\
SQF-stabilized-P-med && 1.243 & 3.2 && 0.569 & -32.9 && 0.670 & 0.4 && 0.073 & -39.7 \\
SQF-stabilized-P-hi  && 1.289 & 7.0 && 0.562 & -33.7 && 0.672 & 0.7 && 0.080 & -33.9 \\
SQF-stabilized-P-max && 1.349 & 12.0 && 0.627 & -26.1 && 0.675 & 1.2 && 0.103 & -14.9 \\
\hdashline
SQF-stabilized-F-lo  && 1.215 & 0.8  && 0.661 & -22.1 && 0.668 & 0.1 && 0.089 & -26.4 \\
SQF-stabilized-F-med && 1.252 & 3.9  && 0.483 & -43.0 && 0.671 & 0.6 && 0.061 & -49.6 \\
SQF-stabilized-F-hi  && 1.343 & 11.5 && 0.281 & -66.9 && 0.677 & 1.5 && 0.034 & -71.9 \\
SQF-stabilized-F-max && 1.557 & 29.2 && 0.000 & -100.0 && 0.696 & 4.3 && 0.000 & -100.0 \\
\midrule
snaive-G && 2.598 & 115.6 && 0.015 & -98.2 && 0.887 & 33.0 && 0.174 & 43.8 \\
snaive-B && 2.661 & 120.8 && 0.111 & -86.9 && 0.888 & 33.1 && 0.210 & 73.6 \\
\hdashline
mean-G && 3.929 & 226.1 && 0.204 & -75.9 && 0.794 & 19.0 && 0.004 & -96.7 \\
mean-B && 3.924 & 225.6 && 0.212 & -75.0 && 0.810 & 21.4 && 0.032 & -73.6 \\
\hdashline
ETS-G && 1.302 & 8.0 && 0.957 & 12.9 && 0.700 & 4.9 && 0.151 & 24.8 \\
ETS-B && 1.286 & 6.7 && 0.957 & 12.9 && 0.700 & 4.9 && 0.159 & 31.4 \\
\bottomrule    
\end{tabular}
}
\end{table}

Table~\ref{tab:q1} further quantifies this trade-off 
by reporting percentage differences in sCRPS and s$W_1$ relative to SQF.
For StableSQF, the table highlights specific 
deteriorations in forecast quality (0.5\%, 1\%, 2.5\%, 5\%, and 10\% increases in sCRPS) and the corresponding 
improvements in stability (decreases in s$W_1$),
based on interpolated sCRPS and s$W_1$ values 
from the trained StableSQF models (shown in Figure~\ref{fig:q1}).
The results confirm that moderate stabilization (i.e., using relatively small $\lambda$ values) yields substantial stability gains at minimal cost in forecast quality.
For instance, a 1\% increase in sCRPS comes with a decrease in s$W_1$ of approximately 30\% for the M4 and 65\% for the M5 dataset. 
Stronger stabilization (i.e., using larger $\lambda$ values) further improves stability,
but only at the cost of increasingly severe quality deterioration.

Inducing stability in SQF forecasts thus exhibits diminishing returns: 
initial quality sacrifices lead to substantial stability gains, but further improvements require steep 
quality deterioration.
For high stability levels, SQF-stabilized-F Pareto dominates StableSQF, 
with SQF-stabilized-F-max completely eliminating instability at a
lower quality loss
than StableSQF models with large $\lambda$ values. 
Excessive stabilization during training flattens the forecasts and can severely harm forecast quality, 
whereas the post-processing SQF-stabilized-F-max approach simply retains the first (longest-horizon) SQF forecast---optimized for quality---made for each period.
For low to moderate stability levels, 
StableSQF models slightly Pareto dominate SQF-stabilized-F variants on the M4 dataset, whereas for M5, both approaches yield (near) Pareto-optimal results.

Finally, we can observe that similar increases in sCRPS lead to larger stability improvements for the M5 dataset than for M4.
This may be partly explained by the difference in forecast quality gaps between SQF and SQF-stabilized-F-max.
The increase in sCRPS for SQF-stabilized-F-max is much larger for M4 (+29.2\%) than for M5 (+4.3\%),
indicating that using a shorter forecast horizon provides only limited gains in forecast quality for M5. 
This may reflect either the presence of long-term patterns in the M5 series that can be predicted far ahead, combined with weak short-term dynamics,
or simply a lack of strong patterns in these series.
The latter interpretation is supported by the sCRPS values of the baselines:
the sCRPS values of the mean baselines are closer to the SQF sCRPS value than those of the snaive baselines.
For the M4 dataset, the opposite holds: the snaive sCRPS values are closer to (but substantially higher than) the SQF sCRPS value,
suggesting that combining learned patterns with up-to-date information on local dynamics leads to better forecast quality.
Since inducing stability dampens responsiveness to such 
local dynamics,
the relative deterioration in forecast quality for a given relative stability improvement is larger for the M4 than for the M5 dataset;
equivalently, the stability gains corresponding to a given relative deterioration in forecast quality are smaller, 
as shown in Table~\ref{tab:q1}.

In summary, the results confirm that the StableSQF forecaster can effectively enhance the stability of distribution-free probabilistic forecasts,
with substantial stability gains achievable at minimal cost in forecast quality.

\subsubsection{Impact of quantile-weighted Wasserstein distances to induce stability}
\label{sec:results_Q2}

\begin{figure}[t!]
\centering
\begin{subfigure}{\textwidth}
    \centering
    \includegraphics[width=\textwidth]{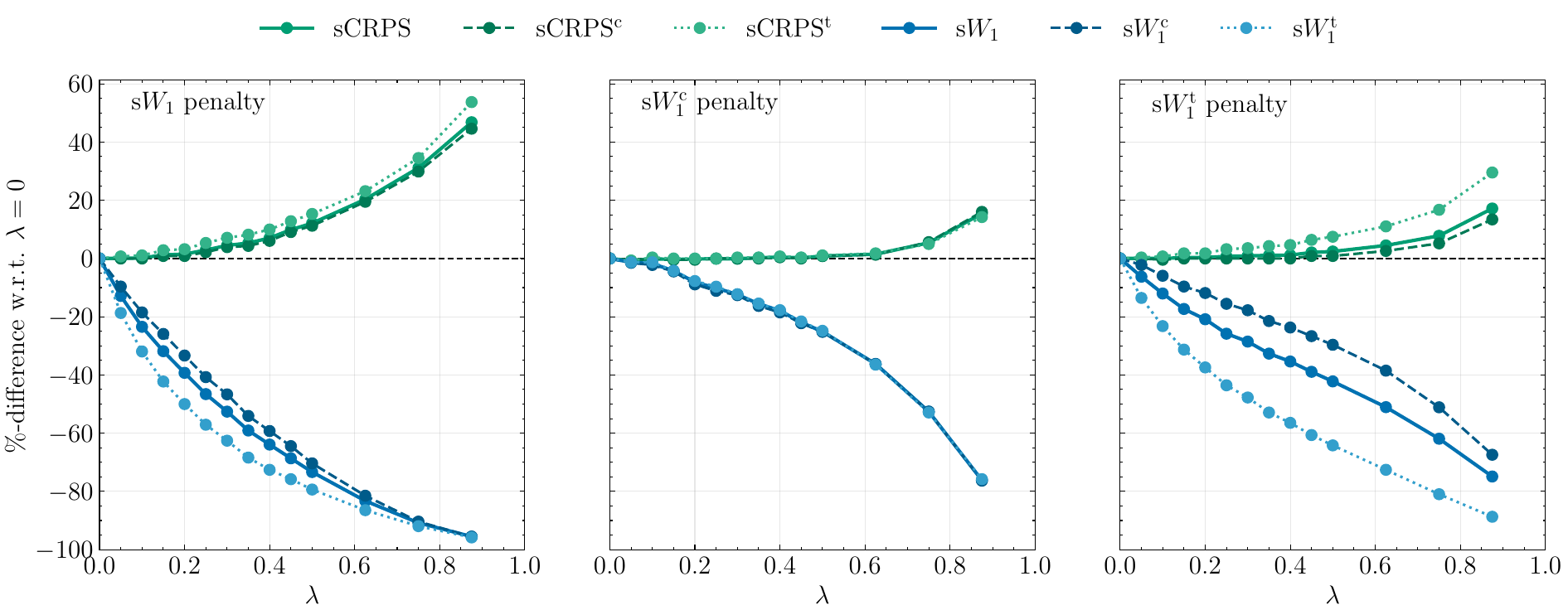}
    \caption{M4 monthly}
    \label{fig:q2_m4}
\end{subfigure}
\vskip 0.8em
\begin{subfigure}{\textwidth}
    \centering
    \includegraphics[width=\textwidth]{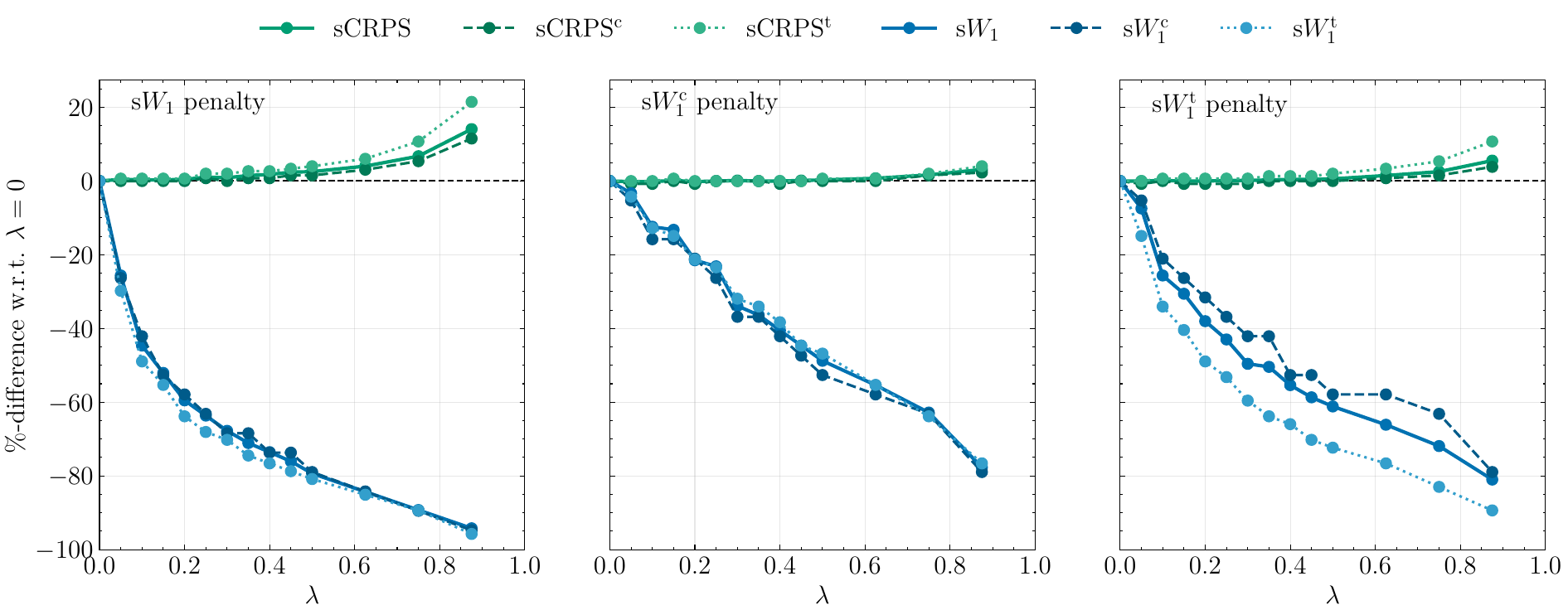}
    \caption{M5 items}
    \label{fig:q2_m5}
\end{subfigure}
\caption{Percentage differences in (quantile-weighted) forecast quality 
(sCRPS, sCRPS$^{\text{c}}$, and sCRPS$^{\text{t}}$)
and 
stability
(s$W_1$, s$W_1^\text{c}$, and s$W_1^\text{t}$)
relative to SQF (i.e., $\lambda = 0$)
for StableSQF variants with different (quantile-weighted) instability penalties in the loss function
(s$W_1$ (left), s$W_1^\text{c}$ (middle), and s$W_1^\text{t}$ (right)). 
Lower values indicate better performance for all metrics.
Results are shown for 
varying penalty weights $\lambda \in [0,1]$.}
\label{fig:q2}
\end{figure}

Figures~\ref{fig:q2_m4} and \ref{fig:q2_m5} visualize the percentage differences in forecast quality and stability metrics for StableSQF variants across penalty weights $\lambda \in [0,1]$ relative to SQF,
for the M4 monthly and M5 items datasets respectively,
for different quantile-weighted instability penalties
used during model optimization
(s$W_1$,
s$W_1^\text{c}$,
and s$W_1^\text{t}$).
For both forecast quality and stability, the uniform quantile-weighted metrics sCRPS and s$W_1$, as well as the center- and tail-focused metrics sCRPS$^\text{c}$, s$W_1^\text{c}$, sCRPS$^\text{t}$, and s$W_1^\text{t}$ (see Section~\ref{sec:evaluation}), are reported.
These figures confirm the findings from the previous section and generalize them to 
different quantile-weighted instability penalties 
as well as to 
quantile-weighted forecast quality and stability metrics.

The key insight from these figures, however, is that different quantile-weighted instability penalties effectively allow stabilization to be targeted toward specific parts of the forecast distributions.
Comparing the percentage differences in the various quality and stability metrics across the different penalties shows that the s$W_1$ penalty
has a slightly stronger stabilizing effect on tail forecasts than on forecasts for the central part of the distributions,
resulting in slightly larger losses in tail-focused than in center-focused forecast quality.
A plausible explanation for this is that tail forecasts are naturally more unstable than 
forecasts for the
central part
(see also Section~\ref{sec:StableSQF}),
leading to larger penalties from the tail quantiles during training since s$W_1$ weights all quantiles uniformly.
The s$W_1^\text{t}$ penalty amplifies this behavior.
In contrast, the s$W_1^\text{c}$ penalty yields more evenly distributed 
relative stability gains and quality losses across the center and tails 
of the forecast distributions.

\begin{table}[t!]
\centering
\caption{Percentage differences in (quantile-weighted) forecast quality 
(sCRPS, sCRPS$^{\text{c}}$, and sCRPS$^{\text{t}}$)
and 
stability 
(s$W_1$, s$W_1^\text{c}$, and s$W_1^\text{t}$)
relative to SQF (i.e., $\lambda = 0$)
for StableSQF variants with different (quantile-weighted) instability penalties 
in the loss function 
(s$W_1$, s$W_1^\text{c}$, and s$W_1^\text{t}$)
and 
SQF-stabilized-F.
Lower values indicate better performance for all metrics.
Percentage differences are calculated using interpolated metric values;
differences that cannot be calculated
are
indicated with `--' (no extrapolation is used).
For a given percentage increase in sCRPS, the best-performing SQF variant per metric is shown in bold and the second best in italics.
Bold and italic formatting is omitted if percentage differences cannot be calculated for all SQF variants.}
\vspace{-0.2cm}
\label{tab:q2}
\resizebox{\textwidth}{!}{
\begin{tabular}{lc c ccccc c ccccc}
\toprule
 &&& \multicolumn{5}{c}{\textbf{M4 monthly}} && \multicolumn{5}{c}{\textbf{M5 items}} \\
 \cmidrule(lr){4-8} \cmidrule(lr){10-14}
 & \textbf{sCRPS} 
 && \textbf{s$\bm{W_1}$} & \textbf{sCRPS$^{\text{c}}$} & \textbf{s$\bm{W_1^\text{c}}$} & \textbf{sCRPS$^\text{t}$} & \textbf{s$\bm{W_1^\text{t}}$}
 && \textbf{s$\bm{W_1}$} & \textbf{sCRPS$^{\text{c}}$} & \textbf{s$\bm{W_1^\text{c}}$} & \textbf{sCRPS$^\text{t}$} & \textbf{s$\bm{W_1^\text{t}}$} \\
\midrule
\multirow{5}{*}{\rotatebox[origin=c]{90}{s$W_1$ penalty}} 
  & \cellcolor{lightgray} 0.5  
  && \textbf{-26.1} & \textit{0.3} & \textbf{-20.8} & 1.6 & \textit{-35.1}
  && \textbf{-60.4} & \textbf{0.0} & \textbf{-59.1} & 0.8 & \textit{-64.5} \\
  & \cellcolor{lightgray} 1.0  
  && \textit{-29.9} & \textit{0.7} & \textit{-24.2} & 2.4 & \textit{-39.9}
  && \textbf{-65.0} & 0.5 & \textbf{-64.9} & 2.0 & \textit{-68.8} \\
  & \cellcolor{lightgray} 2.5  
  && \textbf{-44.5} & \textit{1.8} & \textit{-38.6} & 4.7 & \textit{-55.1}
  && \textit{-78.3} & \textbf{1.5} & \textit{-77.2} & \textit{3.8} & -80.2 \\
  & \cellcolor{lightgray} 5.0  
  && \textbf{-56.6} & \textit{4.2} & \textbf{-51.3} & 7.8 & \textit{-66.2}
  && -86.0 & 3.9 & -86.1 & 7.7 & -86.6 \\
  & \cellcolor{lightgray} 10.0 
  && \textbf{-68.9} & \textit{9.2} & \textbf{-64.8} & 12.9 & \textit{-76.0}
  && -91.5 & 8.1 & -91.8 & 15.5 & -92.2 \\
\midrule
\multirow{5}{*}{\rotatebox[origin=c]{90}{s$W_1^\text{c}$ penalty}} 
  & \cellcolor{lightgray} 0.5  
  && -18.0 & 0.4 & \textit{-18.6} & \textit{0.7} & -17.8
  && -51.7 & \textbf{0.0} & \textit{-55.0} & \textit{0.7} & -50.6 \\
  & \cellcolor{lightgray} 1.0  
  && -28.0 & 1.0 & \textbf{-28.0} & \textit{1.2} & -27.8
  && -57.4 & \textit{0.4} & -59.4 & \textbf{1.0} & -57.7 \\
  & \cellcolor{lightgray} 2.5  
  && -40.4 & 2.4 & \textbf{-40.5} & \textit{2.6} & -40.7
  && -71.3 & 2.0 & -72.1 & \textbf{3.2} & -71.1 \\
  & \cellcolor{lightgray} 5.0  
  && -50.9 & 5.2 & \textit{-50.9} & \textit{4.7} & -51.2
  && -- & -- & -- & -- & -- \\
  & \cellcolor{lightgray} 10.0 
  && -63.4 & 10.4 & \textit{-63.5} & \textbf{9.2} & -63.4
  && -- & -- & -- & -- & -- \\
\midrule
\multirow{5}{*}{\rotatebox[origin=c]{90}{s$W_1^\text{t}$ penalty}} 
  & \cellcolor{lightgray} 0.5 
  && \textit{-22.9} & \textbf{0.0} & -13.4 & 2.4 & \textbf{-39.9}
  && \textit{-59.5} & \textbf{0.0} & -54.4 & 1.6 & \textbf{-70.9} \\
  & \cellcolor{lightgray} 1.0  
  && \textbf{-30.7} & \textbf{0.0} & -19.7 & 3.9 & \textbf{-50.5}
  && \textit{-63.4} & \textbf{0.3} & -57.9 & 2.6 & \textbf{-74.2} \\
  & \cellcolor{lightgray} 2.5  
  && \textit{-42.6} & \textbf{0.9} & -30.0 & 7.6 & \textbf{-64.6}
  && -71.6 & \textbf{1.5} & -62.9 & 5.3 & \textbf{-82.7} \\
  & \cellcolor{lightgray} 5.0  
  && \textit{-52.8} & \textbf{3.0} & -40.5 & 11.9 & \textbf{-73.9}
  && -79.3 & 3.4 & -76.1 & 9.8 & -88.2 \\
  & \cellcolor{lightgray} 10.0 
  && \textit{-65.0} & \textbf{7.1} & -54.9 & 19.7 & \textbf{-82.8}
  && -- & -- & -- & -- & -- \\
\midrule
\midrule
\multirow{5}{*}{\rotatebox[origin=c]{90}{Stabilized-F}} 
  & \cellcolor{lightgray} 0.5 
  && -13.3 & 0.5 & -13.8 & \textbf{0.4} & -12.8
  && -44.5 & \textbf{0.0} & -46.8 & \textbf{0.5} & -45.9 \\
  & \cellcolor{lightgray} 1.0  
  && -23.2 & 1.0 & -24.1 & \textbf{0.9} & -22.4
  && -59.5 & \textbf{0.3} & \textit{-62.0} & \textit{1.3} & -60.5 \\
  & \cellcolor{lightgray} 2.5  
  && -33.5 & 2.5 & -34.2 & \textbf{2.3} & -32.5
  && \textbf{-81.8} & \textit{1.9} & \textbf{-82.9} & \textbf{3.2} & \textit{-82.1} \\
  & \cellcolor{lightgray} 5.0  
  && -46.5 & 5.0 & -47.2 & \textbf{4.6} & -45.4
  && -- & -- & -- & -- & -- \\
  & \cellcolor{lightgray} 10.0 
  && -62.3 & 10.2 & -62.8 & \textit{9.6} & -61.2
  && -- & -- & -- & -- & -- \\
\bottomrule
\end{tabular}
}
\end{table}

Table~\ref{tab:q2} reports the percentage differences visualized in Figure~\ref{fig:q2} for the same specific increases in 
sCRPS (relative to SQF) as in Table~\ref{tab:q1}.
When comparing the percentage differences per metric across the different quantile-weighted instability penalties,
we observe the largest percentage decreases in s$W_1$ for instability penalty s$W_1$.
For s$W_1^\text{t}$,
the largest percentage decreases are observed for instability penalty s$W_1^\text{t}$, 
which comes at the cost of the largest tail-focused quality losses (sCRPS$^\text{t}$)
and the smallest impact on center-focused quality and stability.
Hence, the instability penalty s$W_1^\text{t}$ effectively targets stabilization toward the tails of the forecast distributions.
For s$W_1^\text{c}$, the overall picture is less clear,
with the largest percentage decreases observed for either the s$W_1$ or s$W_1^\text{c}$ penalty
(at the cost of the largest percentage increases in sCRPS$^\text{c}$).
Nevertheless, 
the s$W_1^\text{c}$ penalty effectively targets stabilization toward the center of the forecast distributions,
as it yields the smallest percentage increases in sCRPS$^\text{t}$
and the smallest percentage decreases in s$W_1^\text{t}$,
thus limiting the stabilization of tail forecasts relative to using s$W_1$ as the instability penalty.

Examining the percentage differences per penalty across the different quality and stability metrics further illustrates these findings.
With s$W_1$ as the instability penalty,
the percentage increase in sCRPS$^\text{c}$/sCRPS$^\text{t}$ is consistently smaller/larger than the corresponding increase in sCRPS, 
while the percentage decrease in s$W_1^\text{c}$/s$W_1^\text{t}$ is generally smaller/larger than the corresponding decrease in s$W_1$. 
These asymmetries also appear when using s$W_1^\text{t}$ as the penalty, with the differences amplified.
In contrast, with s$W_1^\text{c}$ as the penalty, the percentage changes are more similar across the uniform quantile-weighted, center-focused, and tail-focused metrics for both quality and stability.

Finally, Table~\ref{tab:q2} also shows the results for SQF-stabilized-F, 
which was shown to achieve competitive quality--stability trade-offs in terms of sCRPS and s$W_1$ in the previous section (see Figure~\ref{fig:q1}).
Its percentage differences indicate that this post-processing approach behaves similarly to using StableSQF with s$W_1^\text{c}$ as the instability penalty, 
in that it does not prioritize stabilizing any particular part of the forecast distributions.

In summary, the results confirm that 
different quantile-weighted 
instability penalties
effectively enable stabilization to be targeted toward specific parts of the forecast distributions,
with the effects on the quality and stability metrics being largest for those aligned with each penalty's focus.

\section{Conclusions}
\label{sec:conclusion}

In this paper, we focused on tackling rolling origin forecast instability in a probabilistic 
setting.
This form of instability arises when time-series forecasts are updated over time, 
and mitigating it is desirable when it may lead to costly adjustments to plans that rely on the forecasts and/or erode trust in the forecasting system.
Specifically, we proposed a method to generate distribution-free probabilistic forecasts based on a parameterized approximation of the quantile function, with the parameters learned by a neural network.
The optimization of this 
network can be directly 
extended to make the resulting forecasts inherently more stable at the cost of a certain reduction in forecast quality,
with the relative importance of stability versus quality being controllable by the modeler.
The method and its extension are referred to as SQF and StableSQF, respectively,
and are generally applicable as no distributional assumptions are made about the forecast distributions.

Our experiments demonstrated that, compared to SQF, StableSQF can yield substantial gains in forecast stability at the cost of only a marginal deterioration in forecast quality.
Further stability improvements are possible by 
placing more weight on instability penalization during 
training,
but come at the cost of increasingly severe quality deterioration.
The appropriate importance weight 
is 
context-dependent,
as it depends on the decisions supported by the forecasts \citep{athanasopoulos2023evaluation}
and the costs associated with forecast quality deterioration and 
forecast 
instability.
As these costs are generally hard to quantify in practice,
and since forecast updating implicitly assumes that the costs of instability are outweighed by the benefits of improved quality,
the goal will often be to obtain a model that generates more stable forecasts without substantial quality deterioration.
If completely eliminating instability is desirable,
applying an extended version of a post-processing approach proposed by \citet{godahewa2025stability} to SQF forecasts 
achieves this at a lower quality loss than StableSQF
and generally yields competitive quality--stability trade-offs.
However, StableSQF provides greater flexibility by enabling stabilization to be targeted toward specific parts of the forecast distributions
while ensuring that the resulting distributions are mathematically valid.
Beyond this flexibility,
the built-in stabilization mechanism reduces operational complexity
by avoiding
the need for an additional post-processing 
step in the forecasting pipeline.

To target instability penalization toward specific parts of the forecast distributions,
we rely on the (quantile) weight functions 
used by 
\citet{gneiting2011} to emphasize either the center or the tails.
However, our experiments showed that the center-focused weight function prevents disproportionate stabilization of the tails rather than emphasizing the center.
Exploring alternative weight functions that place greater emphasis on the center or focus 
on one tail only \citep[see, e.g.,][for weight functions that focus on only the right or left tail]{gneiting2011} and 
selecting
the most appropriate function depending on the intended downstream use 
are promising directions for future research.
In line with this, assessing how more stable probabilistic forecasts affect downstream decision-making performance would be valuable, 
though often challenging due to the difficulty of 
quantifying the costs of adjusting operational plans and the potential impact of lost trust.
Finally, 
the architectural backbone of 
SQF 
can be 
replaced by other direct multi-horizon
architectures that 
could be extended to generate spline-based conditional quantile functions,
such as N-BEATSx \citep{olivares2023nbeatsx} and N-HiTS \citep{challu2023nhits},
to leverage 
exogenous 
covariates and 
the strength of non-overlapping temporal aggregation, respectively.
Investigating their impact on the quality--stability trade-off is another valuable avenue for future work.

\appendix

\section*{Acknowledgements}

Jente Van Belle gratefully acknowledges the financial support from Research Foundation Flanders [grant number 12AZX24N].

\section*{CRediT authorship contribution statement}

\textbf{Jente Van Belle}: Conceptualization, Funding acquisition, Investigation, Methodology, Project administration, 
Software, Visualization, Writing - original draft, Writing - review \& editing.
\textbf{Honglin Wen}: Methodology, Supervision, Writing - review \& editing.
\textbf{Wouter Verbeke}: Funding acquisition, Supervision, Writing - review \& editing.
\textbf{Pierre Pinson}: Conceptualization, Methodology, Supervision, Writing - review \& editing.

\section*{Declaration of use of generative AI and AI-assisted tools in manuscript preparation}

During the preparation of this work, the authors used ChatGPT and Claude to improve language and readability. After using these tools, the authors reviewed and edited the content as needed and take full responsibility for the content of the 
article.

\section*{Declaration of competing interest}

The authors declare that they have no known competing financial interests or personal relationships that could have appeared to influence the work reported in this paper.

\bibliographystyle{elsarticle-harv}
\bibliography{bibliography}

\end{document}